\documentclass{article}

\PassOptionsToPackage{numbers, compress}{natbib}

\usepackage{iclr2027_conference,times}

\usepackage[utf8]{inputenc}
\usepackage[T1]{fontenc}
\usepackage{hyperref}
\usepackage{url}
\usepackage{booktabs}
\usepackage{amsfonts}
\usepackage{microtype}
\usepackage{xcolor}
\usepackage{inconsolata}
\usepackage{amsmath}
\usepackage{graphicx}
\usepackage{caption}
\usepackage{wrapfig}
\usepackage{amsthm}
\theoremstyle{plain}
\newtheorem{proposition}{Proposition}[section]
\newtheorem{corollary}[proposition]{Corollary}
\definecolor{better}{RGB}{30,120,60}
\definecolor{worse}{RGB}{185,45,45}
\title{SpecValley: Aggregation as Spectral Valley in Dense Heterophilic Graphs}

\author{%
\normalfont
\begin{tabular}[t]{@{}p{0.46\textwidth}p{0.46\textwidth}@{}}
\textbf{Christoph Jahn}
  & \textbf{Delvin Bacho} \\
Mercedes-Benz AG, Sindelfingen    & Mercedes-Benz AG, Sindelfingen \\
Department of Computer Science
\\ University of Konstanz            & \\
\texttt{christoph.jahn@mercedes-benz.com}  & \\[2ex]
\textbf{Urs Waldmann}                & \textbf{Bastian Goldlücke} \\
Department of Computer Science        & Department of Computer Science \\
University of Konstanz            & University of Konstanz \\
\texttt{urs.wladmann@uni-konstanz.de}    & \texttt{bastian.goldluecke@uni-konstanz.de}
\end{tabular}}

\iclrfinalcopy

\begin{document}

\maketitle
\fancyhead{}
\fancyhead[L]{Preprint}
\renewcommand{\headrulewidth}{0pt}

\begin{abstract}
Node classification in heterophilic graphs remains an open problem: aggregating
over neighbourhoods can damp precisely the part of the signal that carries the
labels. The field has answered with ever more expressive propagation operators,
leaving the aggregation step that precedes them fixed by degree normalisation.
However, degree normalisation acts as a spectral valley
in its own right, attenuating a fixed middle band whose position at
$\lambda = 1$ is set by the normalisation rather than by the data, exactly where
the eigenvalues of dense graphs accumulate.
We therefore begin with the representations rather than the propagation, in two
stages: a learnable aggregation followed by an ordinary spectral filter. The
first, SpecValley learns the position, sharpness and flank gains of the valley
for every graph.
We show formally how the aggregation operator and downstream filter interact. We
establish that the response $1-\lambda$ of degree normalisation vanishes exactly
at that accumulation point. A spectral energy budget shows that what
aggregation removes from a band, no linear spectral filter can recover.
Crowding alone says nothing about the benefit. That is decided by how much label
energy the valley removes, which we measure for every graph.
SpecValley achieves strong results across heterophilic benchmarks, and a
controlled ablation over aggregation operators and downstream filters attributes
the gain to aggregation rather than the filter.
\end{abstract}

\section{Introduction}
\label{sec:introduction}
Graphs are a fundamental data structure for relational information: nodes stand
for entities, edges for the relations between them. Node classification is a
central task, with applications in social networks, biological systems and
recommendation settings~\citep{kipf2017semi, velickovic2018graph}. Graph neural
networks (GNNs) shape this field by propagating and aggregating information
across local neighbourhoods.

Most GNNs assume homophily, that is, that neighbouring nodes share similar
features or classes~\citep{zhu2020beyond, pei2020geomgcn}. In heterophilic
graphs this does not hold~\citep{yan2021ggcn}: repeated aggregation amplifies
mismatched signals and reduces class separability, particularly in dense
graphs~\citep{oono2020graph}. The field has answered with feature-driven
models~\citep{abu2019mixhop, lim2021linkx}, contrastive
learning~\citep{velickovic2019deep, hassani2020contrastive, qiu2020gcc},
structure learning~\citep{franceschi2019lds, wu2022kdga} and increasingly
elaborate spectral and transformer architectures~\citep{chien2021gprgnn,
lin2024equivariant, Koke2024, deng2024polynormer}, each tied to a regime:
transformers require large datasets, spectral models excel on small or dense
heterophilic graphs. All of them design the convolution operator or the
structure of the graph and leave the step before it untouched: the filter is
learned, the material it operates on is not.

That step is not neutral. Degree normalisation acts spectrally as a fixed valley
with its zero at $\lambda = 1$, a position set by the normalisation rather than
by the data, and it is exactly there that the eigenvalues of dense graphs
accumulate. Whether this helps depends not on the crowding itself, which every
dense graph shows, but on how much label energy lies in the attenuated band.
We therefore treat aggregation as a stage of its own, upstream of an ordinary
spectral filter and make it learnable: SpecValley learns the position,
sharpness and flank gains of the valley per graph, staying close to degree
normalisation where the band carries little label energy and moving its centre
away from $\lambda = 1$ where it carries much.

\textbf{Contributions.}
We contribute a learnable aggregation stage that generalises degree normalisation and
fixes the spectral energy budget available to every downstream filter.
A formal account of the interaction shows that the response $1-\lambda$ vanishes at
the accumulation point of dense spectra, and no linear spectral filter can
recover what aggregation has removed from a band.
We perform a controlled study on stored splits, crossing five aggregators with four
filters on six graphs, together with the label-energy spectrum of each graph
and the response SpecValley learns on it. SpecValley achieves strong results
throughout. The ablation attributes the gain to the aggregation stage rather
than to the filter and learned attention adds little beyond a fixed
aggregation structure once a spectral filter follows it
(Section~\ref{sec:ablations}).

\begin{figure}[t]
    \centering
    \includegraphics[width=1.0\linewidth]{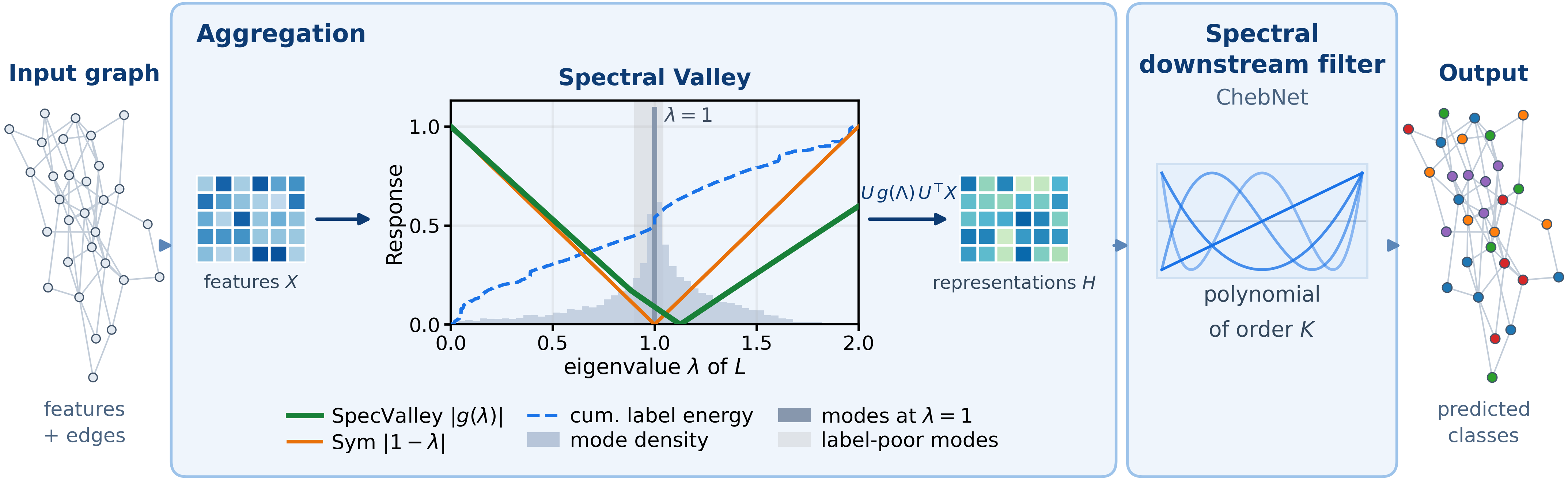}
    \caption{Overview of SpecValley. Top: node features first pass an
    aggregation stage and then an ordinary spectral filter
    (ChebNet, Eq.~\eqref{eq:cheb_conv}), which predicts the node labels.
    Bottom: A more detailed visualization of the aggregation stage. On dense graphs the eigenvalues of
    the normalised Laplacian accumulate at $\lambda = 1$ (grey histogram) and
    degree normalisation acts there as the fixed response $1-\lambda$, a valley
    whose zero is pinned by the normalisation rather than by the data
    (Proposition~\ref{prop:dense_valley}). SpecValley replaces it by a response
    whose centre, sharpness and flank gains are learned per graph
    (Eq.~\eqref{eq:valley_operator}), so that the attenuated band can be placed
    away from the modes that carry the label energy.}
    \label{fig:SpecValley_overview}
\end{figure}

\section{Related Work}
\label{sec:related_work}
Standard neighbourhood aggregation amplifies mismatched signals in heterophilic
graphs, causing oversmoothing and reduced class
separability~\citep{li2018deeper, yan2021ggcn, oono2020graph}.
Feature-driven methods emphasise multi-hop interactions or decouple features
from topology~\citep{abu2019mixhop, lim2021linkx},
structure-oriented methods reweight or denoise edges~\citep{franceschi2019lds,
wu2022kdga}. Both operate on fixed representations.
Spectral methods such as GPR-GNN~\citep{chien2021gprgnn} and
FAGCN~\citep{bo2021fagcn} learn adaptive frequency weights,
BernNet~\citep{he2021bernnet} approximates arbitrary filters,
Specformer~\citep{bo2023specformer} learns the filter with a transformer over
the eigenvalues, JacobiConv~\citep{wang2022jacobiconv} uses Jacobi polynomial
bases, NLSF~\citep{lin2024equivariant} introduces equivariant nonlinear
filtering. These design expressive operators but apply them to a frequency
profile that the preceding aggregation has already fixed.
Few works address that step itself, and they do so spatially: separating ego
from neighbourhood information~\citep{zhu2020beyond, platonov2023critical},
projecting along and against the edge direction~\citep{rossi2023dirgnn} or
learning edge weights by attention~\citep{velickovic2018graph}. Read spectrally,
ego-neighbour separation yields the affine family $a + b\,(1-\lambda)$, so the
position of the response is learnable but its shape is not: sharpness and
asymmetry remain those of degree normalisation. Attention weights are not a function of $L$, and our ablation measures how they compare with degree normalisation (Section~\ref{sec:ablations}).
Graph transformers inject global structure at the feature level via Laplacian or
random-walk encodings~\citep{dwivedi2022lspe, rampasek2022gps,
kreuzer2021rethinking} and Polynormer~\citep{deng2024polynormer} combines polynomial expressivity with local and global attention.

\textbf{Our Positioning.}
We treat the aggregation stage itself as a spectral operator and make it
learnable, with degree normalisation as the special case of a fixed valley whose
position, sharpness and flanks we instead learn. Since aggregation fixes how
much spectral energy reaches the downstream filters at all, our approach does
not replace filter design but precedes it.

\section{Spectral Representations}
\label{sec:method}
\subsection{Problem Formulation}
\label{subsec:problem_formulation}
Let $G=(V,E)$ be a graph on $n$ nodes with features $x_i \in \mathbb{R}^d$,
collected in $X \in \mathbb{R}^{n \times d}$, and adjacency
$A \in \mathbb{R}^{n \times n}$. We work in the transductive setting, where $A$
is fixed and the task is to learn node representations
$Z \in \mathbb{R}^{n \times d'}$ for (semi-)supervised node classification.
Message-passing GNNs obtain them by repeatedly aggregating each node's neighbourhood and combining the result with the node's own state (Gilmer et al., 2017). This paper concerns the aggregation operator, not the propagation that follows.

\subsection{Linear Aggregators with and without Parameters}
\label{subsec:attention_representations}

In heterophilic graphs adjacent nodes are often only weakly aligned
semantically, which makes uniform aggregation
ineffective~\citep{dai2021labelwise}. We therefore view the
encoder as an aggregation operator $P \in \mathbb{R}^{n \times n}$ supported on
$E$ that refines node representations before graph convolution, $H' = P\,V$,
with learnable value projection $V = XW_V$.

\textbf{Parameter-free aggregators.}
As references we use two degree-based operators, the symmetric normalisation of
GCN~\citep{kipf2017semi, wu2019sgc} and mean aggregation as in
GraphSAGE~\citep{hamilton2017inductive},
\begin{equation}
\textstyle
P_{\mathrm{sym}} = D^{-\frac12}AD^{-\frac12}\!:\, h_i' = \sum_{j \in \mathcal{N}(i)} \frac{V_j}{\sqrt{d_i d_j}},
\;\;
P_{\mathrm{mean}} = D^{-1}A\!:\, h_i' = \frac{1}{d_i}\sum_{j \in \mathcal{N}(i)} V_j .
\label{eq:linear_aggregators}
\end{equation}
Both are linear and fixed by the graph; only $W_V$ is learned. For undirected
$A$, $P_{\mathrm{sym}} = I - L$ shares the eigenbasis of the normalised
Laplacian, whereas the row-stochastic
$P_{\mathrm{mean}} = D^{-1/2} P_{\mathrm{sym}} D^{1/2}$ is similar to it but not
symmetric~\citep{chung1997spectral}.

\textbf{Learned aggregation via self-attention.}
Instead of fixing edge weights by degree, we learn them from the features via
scaled dot-product self-attention, which has shown advantages over uniform
aggregation~\citep{velickovic2018graph, zhao2021gophormer}. With query ($Q$),
key ($K$), and value ($V$) projections of the node features and hidden
dimensionality $F_{\mathrm{hidden}}$,
\begin{equation}
e_{ij} = \frac{Q_i^\top K_j}{\sqrt{F_{\mathrm{hidden}}}}
\hspace{0.5cm}
\alpha_{ij} =
\frac{\exp(e_{ij})}
{\sum_{k \in \mathcal{N}(i)} \exp(e_{ik})}
\hspace{0.5cm}
h_i' = \sum_{j \in \mathcal{N}(i)} \alpha_{ij} V_j
\label{eq:attn_aggregation}
\end{equation}
Here, $e_{ij}$ measures feature compatibility and $\alpha_{ij}$ reweights the
neighbours in $\mathcal{N}(i)$, so that $h_i'$ is the input to the subsequent
spectral layer. In matrix form, $H' = A(X)\,V$ with $A(X)_{ij} = \alpha_{ij}$:
attention keeps the support and, like $P_{\mathrm{mean}}$, the row-stochasticity
of Eq.~\eqref{eq:linear_aggregators}, but its weights depend on the features, so
$A(X)$ is not linear in $X$. With frozen query and key projections
(\emph{attn}\textsubscript{init}, Section~\ref{sec:ablations}), $A(X)$ is a fixed
matrix on the transductive graph and acts linearly on $V$. Although
Eq.~\eqref{eq:attn_aggregation} is a local 1-hop message-passing step, it differs
from Graph Attention Networks (GATs)~\citep{velickovic2018graph} in being applied
once as an input encoder rather than interleaved with multi-hop propagation.

\subsection{Adaptive Spectral Aggregation}
\label{subsec:adaptive_spectral_aggregation}

The fixed aggregators of Eq.~\eqref{eq:linear_aggregators} act spectrally through the
response $1-\lambda$, a valley at $\lambda = 1$ whose position and shape follow
from the normalisation rather than from the data (Section~\ref{sec:theory}). In
combination with ChebNet, learned attention behaves similarly on several graphs (Section~\ref{sec:ablations}).
We therefore propose an aggregator that learns the valley per graph:
\begin{equation}
H' \;=\; \underbrace{X W_I}_{\text{ego}}
\;+\; \underbrace{S X W_S + S^{\top} X W_T}_{\text{neighbours}}
\;+\; \underbrace{R(L)\, X W_R}_{\text{spectral valley}},
\qquad
R(L) = U\, r(\Lambda)\, U^{\top}.
\label{eq:valley_operator}
\end{equation}
Aggregation does not only apply the response $1-\lambda$, it also dilutes each
node's own features. The ego term preserves them, which matters where labels are
node-level, and the neighbour
terms aggregate with $S = D^{-1/2}AD^{-1/2}$ along and $S^{\top}$ against the
edge direction~\citep{rossi2023dirgnn};
on undirected graphs both reduce to
$P_{\mathrm{sym}}$. Separate projections
$W_\bullet \in \mathbb{R}^{d \times F_{\mathrm{hidden}}}$ let the model read ego
and neighbour information differently, as recommended for heterophilic
graphs~\citep{zhu2020beyond, lim2021linkx, platonov2023critical}. The spectral
term rescales each Laplacian eigenmode $\lambda$ by
\begin{equation}
r(\lambda) = a(\lambda)\,|\lambda - c|^{p},
\qquad
a(\lambda) = a_L \;\text{for}\; \lambda < c, \quad a(\lambda) = a_R \;\text{for}\; \lambda \ge c.
\label{eq:valley_response}
\end{equation}
The response vanishes at the valley centre $c \in (0, \lambda_{\max})$, the
exponent $p > 0$ sets its sharpness, narrow for $p < 1$ and broad for $p > 1$,
and the side gains $a_L, a_R \in \mathbb{R}$ set slope and sign of either flank.
With $c = p = 1$, $a_L = a_R = 1$ gives $|1-\lambda|$ and $a_L = -a_R = 1$ the
signed response $1-\lambda$ of $P_{\mathrm{sym}}$, so the fixed aggregators are
special cases; we initialise at $|1-\lambda|$ so that training learns departures
from them. Each term is scaled by a scalar gate $\tau_\bullet$ (initialised to
$1$), which adds no expressivity but is fitted separately: all projections
$W_\bullet$ and the operator parameters $\theta = (c, p, a_L, a_R, \tau)$ are
learned on the training loss, the latter by a separate optimiser without weight
decay, and the epoch is selected on the validation set. Test labels are never
used.

\subsection{Spectral Propagation}
\label{subsec:spectral_propagation}

Building on Sections~\ref{subsec:problem_formulation}
to~\ref{subsec:adaptive_spectral_aggregation}, we employ a spectral graph
convolution to propagate information via the graph Laplacian, whose eigenbasis
captures global frequency structure beyond local
neighborhoods~\citep{bruna2014spectral, defferrard2016chebnet}.
Let $L \in \mathbb{R}^{n \times n}$ denote the normalized Laplacian and
$\tilde{L} := 2L/\lambda_{\max} - I = L - I$ its version shifted to $[-1,1]$, using the bound $\lambda_{\max} = 2$.
For node features
$H^{(l)} \in \mathbb{R}^{n \times d'}$, spectral propagation is defined as
\begin{equation}
H^{(l+1)} = \mathrm{LN}\!\left( H^{(l)} + \mathrm{ReLU}\!\left( \sum_{k=0}^{K-1}
T_k(\tilde{L})\, H^{(l)} W_k \right) \right),
\label{eq:cheb_conv}
\end{equation}
with learnable parameters $W_k$, layer normalization $\mathrm{LN}$, and
Chebyshev polynomials $T_k(\tilde{L})$, which follow the recursion
\begin{equation}
T_0(\tilde{L}) = I, \hspace{0.8cm}
T_1(\tilde{L}) = \tilde{L}, \hspace{0.8cm}
T_k(\tilde{L}) = 2\tilde{L}\,T_{k-1}(\tilde{L}) - T_{k-2}(\tilde{L}), \quad k \ge 2.
\label{eq:cheb_k}
\end{equation}
This Chebyshev convolution, in the style of ChebNet~\citep{defferrard2016chebnet},
approximates spectral filters without eigendecomposition.
Since $|T_k| \le 1$ on $[-1, 1]$, the recursion is numerically stable.
We use a standard Chebyshev filter rather than adaptive spectral operators such as NLSF or HeroFilter~\citep{lin2024equivariant, zhang2025herofilter},
since spectral filters are effective under heterophily, where discriminative signal often sits at high frequencies~\citep{chien2021gprgnn, bo2021fagcn},
and a fixed downstream filter isolates the effect of the aggregation stage.

\textbf{SpecValley} combines $\ell = 3$ such layers of order $K = 5$ with the
adaptive valley aggregator of Eq.~\eqref{eq:valley_operator}, which supplies
$H^{(0)}$; the ablation in Section~\ref{sec:ablations} substitutes the other
aggregators of Section~\ref{subsec:attention_representations} in the same
position. Propagation thus operates on aggregated rather than raw inputs.

\section{Theoretical Properties of Aggregation Representations}
\label{sec:theory}

\subsection{Computational Complexity}
\label{subsec:complexity}

\begin{wraptable}{r}{0.46\textwidth}
\vspace{-5pt}
\centering
\small
\begin{tabular*}{\linewidth}{@{}l@{\extracolsep{\fill}}r@{}}
\toprule
\textbf{Model} & \textbf{Cost per forward pass} \\
\midrule
ChebNet                          & $C_0$ \\
$P_{\mathrm{sym}}$ + ChebNet      & $C_0 + O(sd')$ \\
SpecValley                       & $C_0 + O(n^2 d') + O(n^3)^{\ast}$ \\
\bottomrule
\end{tabular*}
\vspace{-3mm}
\caption{Complexity; $C_0$ covers the input projection and the ChebNet
layers. $^{\ast}$Once per graph.}
\label{tab:complexity}
\end{wraptable}
Let $n$ be the number of nodes, $s = |E|$ the number of edges, and $d$ and $d'$
the input and hidden dimension. Every model projects the features to $d'$ and
applies~$\ell$ polynomial filter layers of order $K$ (ChebNet in SpecValley),
at a common cost
$C_0 := O(ndd') + \ell\, C_K$ with $C_K = O\big(K(nd'^2 + sd')\big)$ per layer
(Table~\ref{tab:complexity}).
$P_{\mathrm{sym}}$ adds one sparse product. The ego and neighbour terms of
SpecValley add $O(ndd' + sd')$; its remaining cost lies in the spectral term
$U r(\Lambda) U^\top X W_R$. The eigendecomposition of $L$ takes $O(n^3)$ time
once per graph, $U$ occupies $O(n^2)$ memory during training, and each pass
multiplies by $U^\top$ and $U$ at $O(n^2 d')$. Learning the valley only
re-evaluates $r$ at the $n$ eigenvalues, at $O(n)$ per step. For
$n > 14{,}000$, a degree-$M$ Chebyshev expansion of $r$ in $L - I$ ($M = 40$)
replaces this path at $O(Msd')$ time and $O(Mnd')$ memory. Bilevel training
adds one forward and backward pass per epoch, less than doubling the training
cost.

\subsection{Uniform Aggregation as a Spectral Valley}
\label{subsec:spectral_denoising}

Analyses of the contextual stochastic block model (CSBM) delimit when
aggregation helps: one graph convolution widens linear separability by a factor
of order $1/\sqrt{\bar d}$ in the expected degree $\bar d$~\citep{baranwal2021graph} and
attention helps only when class means are well separated or structural noise
dominates, collapsing onto uniform convolution
otherwise~\citep{fountoulakis2023attention, javaloy2023learnable,
ma2025attention}. Our datasets are not CSBM graphs, so these results frame
rather than predict our observations; on dense graphs with weak features, they
point to uniform aggregation.

For this, let $A$ be symmetric and nonnegative, without self-loops and isolated
nodes, with degree matrix $D$, and let
$L = I - P_{\mathrm{sym}} = U\Lambda U^\top$ with
$P_{\mathrm{sym}} = D^{-1/2}AD^{-1/2}$. We call
$d_{\mathrm{eff}} := n/\|P_{\mathrm{sym}}\|_F^2$ the \emph{effective degree};
for unweighted $A$ it lies between the minimum and maximum degree and equals $r$
on $r$-regular graphs (Appendix~\ref{app:lemma_valley}).

\begin{proposition}
\label{prop:dense_valley}
In the setting above, $\frac{1}{n}\sum_k \lambda_k = 1$ and
$\frac{1}{n}\sum_k (\lambda_k - 1)^2 = 1/d_{\mathrm{eff}}$, and
$P_{\mathrm{sym}} = U(I-\Lambda)U^\top$ has the spectral response $1-\lambda$,
which vanishes at this mean. If moreover $d_{\mathrm{eff}} \ge 1/(\beta^2\eta)$
for some $\beta > 0$ and $\eta \in (0,1)$, then at least $(1-\eta)\,n$ of the
eigenvalues $\lambda_1,\dots,\lambda_n$ satisfy $|\lambda_k - 1| < \beta$, and
$P_{\mathrm{sym}}$ scales each of the corresponding modes by a factor of
absolute value less than $\beta$.
\end{proposition}

The second statement is Chebyshev's inequality applied to the spectrum and holds
for every graph; for random graphs with minimum expected degree well above
$\log^2 n$, all non-trivial eigenvalues concentrate at
$1$~\citep{chung2003spectra}. Uniform aggregation thus places its zero where
dense spectra accumulate: it is a fixed spectral valley rather than a low-pass
filter, attenuating the bulk and keeping both ends of the spectrum, where
community ($\lambda \approx 0$) and alternating ($\lambda \approx 2$) structure
reside.

\textbf{Interpretation.} Proposition~\ref{prop:dense_valley} explains why
aggregation helps on dense graphs and where it acts. Isotropic noise spreads
evenly over the modes, so on dense graphs most of it lies in the band that
$P_{\mathrm{sym}}$ attenuates, which retains exactly a fraction
$1/d_{\mathrm{eff}}$ of the expected noise energy. Whether the labels survive is
a separate question: label energy in the same band is attenuated along with the
noise. Since no fixed response suits every graph, this motivates learning the
valley (Section~\ref{subsec:adaptive_spectral_aggregation}).

\subsection{The Spectral Energy Budget of Aggregation}
\label{subsec:theory_energy_budget}

The downstream filter works with what aggregation leaves. For this, fix
\emph{a priori} a partition $\{\mathcal{I}_b\}_{b=1}^m$ of the frequency indices
into bands with projections $P_b := \sum_{k \in \mathcal{I}_b} u_k u_k^\top$ and
band energies $E_b(Y) := \|P_b Y\|_F^2$. Let
$\mathcal{G}_\Gamma := \{\, g : \sigma(L) \to \mathbb{R} \,:\, \max_k
|g(\lambda_k)| \le \Gamma \,\}$ be the admissible filters,
$g(L) = U g(\Lambda) U^\top$, with extremal gains
$\mu_b^-(g) := \min_{k \in \mathcal{I}_b} |g(\lambda_k)|$ and
$\mu_b^+(g) := \max_{k \in \mathcal{I}_b} |g(\lambda_k)|$; $g$ is
\emph{band-constant} if $\mu_b^-(g) = \mu_b^+(g)$ for all $b$.

\begin{corollary}
\label{cor:energy_budget}
In the setting above, let $Y$ be a graph signal, such as the aggregated features
$Y = A(x)X$, and $Z := g(L)\,Y$ with $g \in \mathcal{G}_\Gamma$. Then:

\emph{(i) Band-wise budget.}
$\|P_b Z\|_F^2 = \sum_{k \in \mathcal{I}_b} g(\lambda_k)^2 \|u_k^\top Y\|_2^2$,
hence $\big(\mu_b^-(g)\big)^2 E_b(Y) \le \|P_b Z\|_F^2 \le
\big(\mu_b^+(g)\big)^2 E_b(Y)$, with equality for band-constant $g$.

\emph{(ii) Non-creation.} If $E_b(Y) = 0$, then $P_b Z = 0$ for every
$g \in \mathcal{G}_\Gamma$.
\end{corollary}

With Proposition~\ref{prop:dense_valley}, (i) quantifies the valley: for
$Y = P_{\mathrm{sym}}X$ and the central band
$\mathcal{I}_\beta := \{k : |\lambda_k - 1| < \beta\}$ with projection $P_\beta$
and band energy $E_\beta$, $u_k^\top Y = (1-\lambda_k)\,u_k^\top X$ gives
$\|P_\beta Z\|_F^2 \le \Gamma^2\beta^2 E_\beta(X)$ for every admissible $g$.
Spectral aggregators move no energy between bands; non-spectral ones can.

\textbf{Interpretation.} The corollary makes the band-energy profile the
interface between aggregation and filter: a linear spectral filter can only reweight what
aggregation supplies (i) and cannot restore what it removed (ii), and on dense
graphs this concerns the band that holds almost all modes. The ego term of
Eq.~\eqref{eq:valley_operator} keeps every band of $X$ available to the filter,
consistent with the adaptive valley avoiding the collapse of the other
aggregators on Actor and Tolokers. And where the label energy is spread over the
modes like noise, as on Actor, no spectral aggregator can improve the ratio of
label energy to noise. Proofs are given in
Appendices~\ref{app:lemma_valley} and~\ref{app:proposition_energy}.

\subsection{Compatibility of Aggregation and Filter}
\label{subsec:Compatibility_Aggregation_Filter}

In the model, aggregation and filter act in sequence. For spectral functions
$q(L) = U q(\Lambda) U^\top$ and $h(L) = U h(\Lambda) U^\top$, diagonal matrices
commute, so $h(L)\,q(L) = U\,h(\Lambda)\,q(\Lambda)\,U^\top = (hq)(L)
= q(L)\,h(L)$,
and weight matrices, acting from the right, do not interfere. Each Chebyshev
term of the first ChebNet layer (Eq.~\eqref{eq:cheb_conv}) applied to
$q(L)XW$ is thus one spectral filter with response
$T_k(\lambda-1)\,q(\lambda)$~\citep{shuman2013emerging}. For degree
normalisation, all these responses vanish at $\lambda = 1$: the pair is a
ChebNet of one order higher with a fixed root~\citep{wu2019sgc}. On
undirected graphs all branches of SpecValley are spectral as well, with
responses $1$, $1-\lambda$ and $r(\lambda)$; the ego branch carries no root,
and since $r$ is in general not a polynomial, neither are the products.
Attention and $P_{\mathrm{mean}}$ on non-regular graphs are not spectral
functions of $L$ and do not commute with $h(L)$ in general, since a matrix
commutes with $h(L)$ only if it maps each of its eigenspaces into
itself~\citep{horn2012matrix}; their output enters the filter as an input
signal, to which Corollary~\ref{cor:energy_budget} applies. All statements
concern the linear step: ReLU and layer normalisation can move energy between
modes.

\section{Experimental Evaluation}
\subsection{Experimental Settings}
\label{sec:experimental_settings}

\begin{wraptable}{r}{0.57\textwidth}
\vspace{-6pt}
\centering
\small
\setlength{\tabcolsep}{3pt}

\begin{tabular}{lrrrrr}
\hline
Dataset        & Nodes    & Edges    & Features & Classes & Degree \\
\hline
Squirrel-f$^{*}$  & 2.2K  & 94K   & 2.1K  & 5  & 42 \\
Chameleon-f$^{*}$ & 0.9K  & 14K   & 2.3K  & 5  & 31 \\
Actor$^{*}$       & 7.6K  & 30K   & 932   & 5  & 8  \\
\hline
Crocodile$^{*}$   & 11.6K & 360K  & 13.2K  & 5  & 62 \\
Tolokers       & 11.8K & 519K  & 10    & 2  & 88 \\
HM-categories  & 46.6K & 21.5M & 120   & 21 & 923 \\
\hline
\end{tabular}

\caption{Dataset statistics. $^{*}$: 48/32/20 train/val/test split;
Tolokers: 50/25/25; HM-categories: 10/10/80 (GraphLand RL split).
Squirrel-f and Chameleon-f denote the node-deduplicated filtered
variants from~\citep{platonov2023critical}.}
\label{tab:dataset_stats_compact}
\end{wraptable}

We evaluate our method on node classification in dense relational
graphs, using six heterophilic benchmarks: Squirrel-f, Chameleon-f,
Actor, Crocodile~\citep{rozemberczki2021multiscale}, Tolokers~\citep{platonov2023critical}
and HM-categories~\citep{bazhenov2025graphland}
(Table~\ref{tab:dataset_stats_compact}). Squirrel-f and Chameleon-f are the
node-deduplicated variants of~\citet{platonov2023critical}, which remove the
train--test leakage caused by duplicated pages, and are used in place of the
originals for comparison with prior work~\citep{lin2024equivariant}. Crocodile is a
denser Wikipedia network from the same source, whose published target is
continuous and which we discretise into five equally sized classes, as done for
Chameleon and Squirrel by~\citet{pei2020geomgcn} and on this graph
by~\citet{xiao2023graphacl}.
The selection spans more than an order of magnitude in size and degree and
deliberately covers two spectral regimes. On Squirrel-f, Chameleon-f, Crocodile
and HM-categories, neighbourhood aggregation clearly improves over the
feature-only baseline (Table~\ref{tab:ablation_spectral}), so label information is
recoverable by spectral denoising. On Actor and Tolokers it is not: Actor is
sparse with a label spectrum close to that of noise
(Section~\ref{subsec:theory_energy_budget}), while Tolokers, and in part
HM-categories, carry label information at the level of individual nodes, where
models that separate ego from neighbour information and attention-based
architectures perform best~\citep{platonov2023critical, bazhenov2025graphland}.
Covering both regimes lets us attribute gains and failures to the spectral
structure of a graph rather than to the choice of benchmarks.

For Squirrel-f, Chameleon-f, Actor and Crocodile we use 10 random
48\%/32\%/20\% train/validation/test splits~\citep{liu2023beyond, yuan2023muse};
for Tolokers the 10 official 50\%/25\%/25\% splits~\citep{platonov2023critical};
and for HM-categories the official GraphLand RL split (10\%/10\%/80\%) with three
seeds~\citep{bazhenov2025graphland}. We report 95\% confidence
intervals~\citep{lin2024equivariant}.
All splits are stored and reused across models. Detailed configurations are provided in
Appendix~\ref{app:implementation_details}.

\subsection{Interplay of Aggregation Operator and Downstream Filter}
\label{sec:ablations}

To isolate the contribution of the aggregation operator, we cross five
aggregators with four downstream filters (Table~\ref{tab:ablation_spectral}).
As parameter-free references we use symmetric normalisation (\emph{sym}) and
mean aggregation (\emph{mean}); as parameterised operators, learned attention
(\emph{attn}), initialised attention (\emph{attn}\textsubscript{init}), whose
query and key projections stay fixed after initialisation so that only the value
projection is learned, and the adaptive valley (\emph{valley},
Section~\ref{subsec:adaptive_spectral_aggregation}). The filters cover the
rational (APPNP~\citep{gasteiger_predict_2019}, ARMA~\citep{bianchi2022arma})
and polynomial
(Gegenbauer~\citep{castro2024gegenbauer,goksu2025gegenbauernet},
ChebNet~\citep{defferrard2016chebnet}) families. A row without aggregation
(\emph{none}) and a block without a filter serve as references. All runs follow
Section~\ref{sec:experimental_settings}, train for 150 epochs, and report 95\%
confidence intervals.

\begin{wraptable}{l}{0.60\textwidth}
\vspace{-5pt}
\centering
\renewcommand{\arraystretch}{0.87}
\setlength{\tabcolsep}{3.5pt}
\resizebox{\linewidth}{!}{%
\scriptsize
\begin{tabular}{@{}ll|ccc|ccc@{}}
\toprule
Filt. & Aggr. & Squirrel$^{*}$ & Chameleon$^{*}$ & Actor & Crocodile
  & Tolokers$^{**}$ & HM-cat. \\
\midrule
 APPNP     & none                         & 36.2$\pm$1.4 & 37.6$\pm$2.2 & \underline{36.0$\pm$0.6} & 69.7$\pm$0.7 & 76.9$\pm$1.1 & 37.7$\pm$0.7 \\
           & sym                          & \textcolor{worse}{33.4$\pm$1.1} & \textcolor{better}{42.5$\pm$2.5} & \textcolor{worse}{24.9$\pm$1.0} & \textcolor{better}{75.7$\pm$0.4} & \textcolor{worse}{75.4$\pm$0.7} & \textcolor{better}{53.1$\pm$0.4} \\
           & mean                         & \textcolor{worse}{33.5$\pm$0.8} & \textcolor{better}{42.4$\pm$3.3} & \textcolor{worse}{25.1$\pm$0.7} & \textcolor{better}{\underline{76.5$\pm$0.5}} & \textcolor{worse}{73.4$\pm$0.7} & \textcolor{better}{51.4$\pm$1.4} \\
           & attn\textsubscript{init}     & \textcolor{worse}{33.9$\pm$1.7} & \textcolor{better}{\underline{43.8$\pm$2.8}} & \textcolor{worse}{24.9$\pm$0.6} & \textcolor{better}{76.0$\pm$0.6} & \textcolor{worse}{73.2$\pm$0.8} & \textcolor{better}{53.5$\pm$0.7} \\
           & attn                         & \textcolor{worse}{34.3$\pm$0.8} & \textcolor{better}{42.3$\pm$2.2} & \textcolor{worse}{24.9$\pm$0.6} & \textcolor{better}{75.8$\pm$0.6} & \textcolor{worse}{75.4$\pm$0.8} & \textcolor{better}{54.4$\pm$0.1} \\
           & valley                       & \textcolor{better}{\underline{39.9$\pm$2.0}} & \textcolor{better}{40.4$\pm$3.0} & \textcolor{worse}{34.8$\pm$0.8} & \textcolor{better}{74.0$\pm$0.8} & \textcolor{better}{\underline{80.6$\pm$0.8}} & \textcolor{better}{\underline{59.3$\pm$0.9}} \\
\midrule
 ARMA      & none                         & \underline{36.4$\pm$1.3} & 37.2$\pm$3.5 & \textbf{\underline{36.4$\pm$0.7}} & 69.8$\pm$0.5 & 74.8$\pm$0.9 & 36.5$\pm$0.3 \\
           & sym                          & \textcolor{worse}{33.2$\pm$1.3} & \textcolor{better}{42.6$\pm$2.8} & \textcolor{worse}{25.2$\pm$0.9} & \textcolor{better}{75.8$\pm$0.5} & \textcolor{better}{75.8$\pm$0.7} & \textcolor{better}{52.2$\pm$0.5} \\
           & mean                         & \textcolor{worse}{32.9$\pm$1.6} & \textcolor{better}{42.3$\pm$2.1} & \textcolor{worse}{25.2$\pm$0.7} & \textcolor{better}{\underline{76.3$\pm$0.6}} & \textcolor{worse}{73.2$\pm$0.6} & \textcolor{better}{51.0$\pm$0.5} \\
           & attn\textsubscript{init}     & \textcolor{worse}{34.2$\pm$1.4} & \textcolor{better}{\underline{43.0$\pm$2.7}} & \textcolor{worse}{25.6$\pm$0.9} & \textcolor{better}{76.2$\pm$0.8} & \textcolor{worse}{73.1$\pm$0.8} & \textcolor{better}{52.6$\pm$0.7} \\
           & attn                         & \textcolor{worse}{33.6$\pm$1.4} & \textcolor{better}{41.5$\pm$2.0} & \textcolor{worse}{25.3$\pm$0.7} & \textcolor{better}{\underline{76.3$\pm$0.7}} & \textcolor{better}{75.9$\pm$0.7} & \textcolor{better}{53.6$\pm$0.1} \\
           & valley                       & \textcolor{worse}{35.8$\pm$1.2} & \textcolor{better}{39.7$\pm$2.3} & \textcolor{worse}{35.0$\pm$1.2} & \textcolor{better}{73.7$\pm$1.0} & \textcolor{better}{\underline{80.3$\pm$0.9}} & \textcolor{better}{\underline{57.4$\pm$3.0}} \\
\midrule
 Gegen.    & none                         & 37.8$\pm$1.2 & 34.2$\pm$2.3 & \underline{33.6$\pm$1.1} & 68.1$\pm$0.6 & \underline{81.9$\pm$0.7} & 47.2$\pm$1.4 \\
           & sym                          & \textcolor{better}{40.5$\pm$1.8} & \textcolor{better}{41.6$\pm$2.8} & \textcolor{worse}{26.7$\pm$1.1} & \textcolor{better}{76.1$\pm$0.6} & \textcolor{worse}{80.8$\pm$0.8} & \textcolor{better}{55.3$\pm$0.3} \\
           & mean                         & \textcolor{better}{40.7$\pm$1.2} & \textcolor{better}{42.6$\pm$2.2} & \textcolor{worse}{26.4$\pm$1.2} & \textcolor{better}{76.3$\pm$0.7} & \textcolor{worse}{79.1$\pm$0.6} & \textcolor{better}{52.2$\pm$0.8} \\
           & attn\textsubscript{init}     & \textcolor{better}{39.5$\pm$1.8} & \textcolor{better}{42.9$\pm$2.7} & \textcolor{worse}{27.2$\pm$0.7} & \textcolor{better}{\underline{76.4$\pm$0.8}} & \textcolor{worse}{79.1$\pm$0.6} & \textcolor{better}{54.3$\pm$1.4} \\
           & attn                         & \textcolor{better}{39.6$\pm$2.0} & \textcolor{better}{41.4$\pm$1.5} & \textcolor{worse}{26.8$\pm$1.0} & \textcolor{better}{\underline{76.4$\pm$0.6}} & \textcolor{worse}{79.1$\pm$0.8} & \textcolor{better}{55.6$\pm$0.4} \\
           & valley                       & \textcolor{better}{\underline{42.5$\pm$1.4}} & \textcolor{better}{\underline{44.0$\pm$3.9}} & \textcolor{worse}{33.3$\pm$0.9} & \textcolor{better}{75.0$\pm$1.0} & \underline{81.9$\pm$0.6} & \textcolor{better}{\underline{59.4$\pm$1.1}} \\
\midrule
 ChebNet   & none                         & 37.5$\pm$1.6 & 37.8$\pm$2.5 & 35.4$\pm$0.6 & 72.4$\pm$0.6 & \textbf{\underline{85.2$\pm$0.5}} & 57.1$\pm$0.6 \\
           & sym                          & \textcolor{better}{42.3$\pm$1.8} & \textcolor{better}{43.1$\pm$1.8} & \textcolor{worse}{29.4$\pm$0.8} & \textcolor{better}{76.7$\pm$0.6} & \textcolor{worse}{82.6$\pm$0.8} & \textcolor{better}{60.3$\pm$1.2} \\
           & mean                         & \textcolor{better}{41.7$\pm$1.4} & \textcolor{better}{43.0$\pm$2.5} & \textcolor{worse}{27.7$\pm$0.9} & \textcolor{better}{77.1$\pm$0.6} & \textcolor{worse}{81.7$\pm$0.8} & \textcolor{better}{57.7$\pm$0.3} \\
           & attn\textsubscript{init}     & \textcolor{better}{41.5$\pm$1.9} & \textcolor{better}{44.3$\pm$2.2} & \textcolor{worse}{27.0$\pm$0.8} & \textcolor{better}{\textbf{\underline{77.3$\pm$0.6}}} & \textcolor{worse}{81.5$\pm$0.7} & \textcolor{better}{60.0$\pm$0.3} \\
           & attn                         & \textcolor{better}{41.7$\pm$2.1} & \textcolor{better}{42.4$\pm$3.4} & \textcolor{worse}{27.4$\pm$1.0} & \textcolor{better}{\textbf{\underline{77.3$\pm$0.5}}} & \textcolor{worse}{81.7$\pm$0.7} & \textcolor{better}{61.0$\pm$0.3} \\
           & valley                       & \textcolor{better}{\textbf{\underline{42.7$\pm$2.6}}} & \textcolor{better}{\textbf{\underline{44.5$\pm$1.5}}} & \textcolor{better}{\underline{35.6$\pm$0.8}} & \textcolor{better}{76.5$\pm$0.8} & \textbf{\underline{85.2$\pm$0.5}} & \textcolor{better}{\textbf{\underline{63.2$\pm$0.9}}} \\
\midrule
 No filter & none                         & 33.5$\pm$1.9 & 35.5$\pm$2.0 & \underline{36.2$\pm$0.8} & 68.4$\pm$0.7 & 71.5$\pm$0.5 & 35.8$\pm$0.4 \\
           & sym                          & \textcolor{worse}{33.2$\pm$1.1} & \textcolor{better}{42.4$\pm$2.7} & \textcolor{worse}{25.3$\pm$0.8} & \textcolor{worse}{58.0$\pm$0.8} & \textcolor{better}{72.3$\pm$0.9} & \textcolor{better}{42.4$\pm$1.6} \\
           & mean                         & \textcolor{worse}{33.0$\pm$1.8} & \textcolor{better}{42.3$\pm$3.2} & \textcolor{worse}{25.1$\pm$0.8} & \textcolor{worse}{66.3$\pm$0.7} & \textcolor{worse}{67.2$\pm$1.1} & \textcolor{better}{38.4$\pm$0.2} \\
           & attn\textsubscript{init}     & \textcolor{worse}{33.4$\pm$1.3} & \textcolor{better}{\underline{43.8$\pm$2.5}} & \textcolor{worse}{25.3$\pm$0.8} & \textcolor{worse}{66.8$\pm$0.7} & \textcolor{worse}{67.3$\pm$1.1} & \textcolor{better}{37.0$\pm$0.7} \\
           & attn                         & \textcolor{worse}{32.9$\pm$1.2} & \textcolor{better}{40.5$\pm$2.7} & \textcolor{worse}{25.3$\pm$0.8} & \textcolor{worse}{66.8$\pm$0.7} & \textcolor{better}{73.0$\pm$0.8} & \textcolor{better}{44.6$\pm$0.3} \\
           & valley                       & \textcolor{better}{\underline{37.3$\pm$1.4}} & \textcolor{better}{40.6$\pm$2.0} & \textcolor{worse}{36.1$\pm$0.6} & \textcolor{better}{\underline{76.2$\pm$0.7}} & \textcolor{better}{\underline{77.3$\pm$0.7}} & \textcolor{better}{\underline{48.1$\pm$0.2}} \\
\bottomrule
\end{tabular}
}
\captionsetup{justification=raggedright,singlelinecheck=false}
\vspace{-3mm}
\caption{All combinations of aggregation operator and downstream filter.
Accuracy (\%), mean $\pm$ 95\% CI over 10 splits, HM-categories over 3 seeds.
$^{*}$node-deduplicated variant of~\citet{platonov2023critical};
$^{**}$ROC-AUC. All runs 150 epochs. \textbf{Bold} marks the best value per dataset,
\underline{underline} the best per filter; colour compares each operator with
\emph{none} under the same filter (\textcolor{better}{better},
\textcolor{worse}{worse}).}
\label{tab:ablation_spectral}
\end{wraptable}

The benefit depends on the spectral regime of the graph.
On Chameleon-f, Crocodile and HM-categories every aggregator improves over \emph{none} under each of the four filters (without a filter, only the valley does so on Crocodile).
On Squirrel-f the picture is split: with the polynomial filters
all of them improve, with APPNP and without a filter only the valley does, and
with ARMA none does. On Actor, and for 16 of 20 combinations on Tolokers, the
parameter-free and attention-based aggregators instead fall below \emph{none},
consistent with label energy that is spread like noise or carried by individual
nodes (Sections~\ref{subsec:spectral_denoising}
and~\ref{subsec:theory_energy_budget}). The adaptive valley avoids this
collapse: on Actor it stays within $1.5$ points of \emph{none}, where the others
lose up to $11$, and on Tolokers it matches \emph{none} under the polynomial
filters and exceeds it by $3.7$ to $5.8$ points otherwise.
Comparing \emph{attn} with \emph{attn}\textsubscript{init} without a downstream
filter tests whether the learned weights add a dataset-specific alignment beyond
the fixed aggregation structure. On Tolokers they gain $5.7$ points there, but
at most $2.8$ once a filter is applied and essentially nothing for the
polynomial filters, so the learned reweighting is not aligned with what the
filter does downstream. HM-categories is the complementary case: attention helps
with and without a filter, consistent with it exploiting node-specific as well
as spectral information.

\subsection{Benchmark Results}
\label{sec:benchmark_results}

SpecValley (valley aggregator + ChebNet, Section~\ref{subsec:adaptive_spectral_aggregation}) is level with or above the best published result on four of six benchmarks, within the confidence interval of the best on Tolokers and behind on Actor (Table~\ref{tab:benchmark}).
The benchmark runs use the protocol of Section~\ref{sec:experimental_settings} with 1000 epochs and a polynomial order $K$ chosen per dataset on the validation score (Appendix~\ref{app:implementation_details}); Table~\ref{tab:ablation_spectral} uses 150 epochs and $K=5$ throughout, which explains the differences between the two tables.

On Squirrel-filtered and Chameleon-filtered, spectral and frequency-aware methods constitute the strongest prior work. SpecValley improves over the next best (NLSFs~\citep{lin2024equivariant}) by $2.4$ and $1.3$ points, with FAGCN~\citep{bo2021fagcn} further below. Both margins lie within the confidence intervals.
On Crocodile the listed prior work is self-supervised (GraphACL~\citep{xiao2023graphacl}, DSSL~\citep{xiao2022dssl}, HGRL~\citep{chen2022hgrl}),
\begin{wraptable}{r}{0.55\textwidth}
\vspace{-5pt}
\centering
\scriptsize
\setlength{\tabcolsep}{2.0pt}
\renewcommand{\arraystretch}{0.95}
\begin{tabular}{@{}c|lr@{}l|lr@{}l|lr@{}l@{}}
\toprule
Rank & \multicolumn{3}{c|}{Squirrel$^{*}$} & \multicolumn{3}{c|}{Chameleon$^{*}$} & \multicolumn{3}{c}{Actor} \\
\midrule
1 & \textbf{SpecValley} & \textbf{45.1} & {\tiny\,$\pm$2.1} & \textbf{SpecValley} & \textbf{43.8} & {\tiny\,$\pm$3.1} & MUSE     & 38.6 & {\tiny\,$\pm$1.3} \\
2 & NLSFs           & 42.7          & {\tiny\,$\pm$1.7} & NLSFs           & 42.5          & {\tiny\,$\pm$1.5} & GloGNN   & 37.7 & {\tiny\,$\pm$1.4} \\
3 & FAGCN           & 41.1          & {\tiny\,$\pm$2.3} & FAGCN           & 41.9          & {\tiny\,$\pm$2.7} & GGCN     & 37.5 & {\tiny\,$\pm$1.6} \\
4 & GCN             & 39.5          & {\tiny\,$\pm$1.5} & ResNet+SGC      & 41.0          & {\tiny\,$\pm$4.5} & \textbf{SpecValley} & \textbf{35.7} & {\tiny\,$\pm$0.9} \\
\midrule
Rank & \multicolumn{3}{c|}{Crocodile} & \multicolumn{3}{c|}{Tolokers$^{**}$} & \multicolumn{3}{c}{HM-cat.} \\
\midrule
1 & \textbf{SpecValley} & \textbf{77.2} & {\tiny\,$\pm$0.5} & Polynorm.-r     & 85.9          & {\tiny\,$\pm$0.7} & \textbf{SpecValley} & \textbf{69.7} & {\tiny\,$\pm$1.2} \\
2 & GraphACL        & 66.2          & {\tiny\,$\pm$0.2} & \textbf{SpecValley} & \textbf{85.2} & {\tiny\,$\pm$0.6} & GT       & 69.2 & {\tiny\,$\pm$0.5} \\
3 & DSSL            & 63.0          & {\tiny\,$\pm$0.5} & GAT-sep         & 83.8          & {\tiny\,$\pm$0.4} & GAT      & 68.0 & {\tiny\,$\pm$0.3} \\
4 & HGRL            & 61.9          & {\tiny\,$\pm$0.5} & Exphormer       & 83.8          & {\tiny\,$\pm$0.8} & GCN      & 61.7 & {\tiny\,$\pm$0.4} \\
\bottomrule
\end{tabular}
\captionsetup{justification=raggedright,singlelinecheck=false}
\vspace{-3mm}
\caption{SpecValley vs. the
three strongest reported baselines per dataset as published. Accuracy (\%), mean $\pm$ 95\%
CI over 10 splits, HM-categories over 3 seeds. $^{**}$ROC-AUC, $^{*}$node-deduplicated~\citep{platonov2023critical}, over 1k epochs. On Crocodile the baselines are self-supervised, so the results are
indicative rather than strictly comparable.}
\label{tab:benchmark}
\end{wraptable}
so we do not read the difference as a margin. On Tolokers and HM-categories, attention-based transformers (Polynormer~\citep{deng2024polynormer}, GAT-sep~\citep{platonov2023critical}, Exphormer~\citep{shirzad2023exphormer}, GT and GAT~\citep{bazhenov2025graphland}) are the strongest baselines. SpecValley leads on HM-categories by $0.5$ points and trails Polynormer-r on Tolokers by $0.7$ points, in both cases within the confidence intervals.
Actor is the dataset where SpecValley is clearly behind: contrastive and graph-structure-aware methods (MUSE~\citep{yuan2023muse}, GloGNN~\citep{li2022finding}, GGCN~\citep{yan2021ggcn}) dominate, consistent with the analysis showing that no spectral aggregator can separate signal from noise when the label energy is spread uniformly across the spectrum (Section~\ref{subsec:spectral_denoising}).

\section{Understanding Spectral Representations}
\label{Understanding_Spectral_Attention_Representations}
\label{sec:Understanding_Preconditioning}

\subsection{Spectral Valley Aggregation}
\label{subsec:spectral_properties_datasets}

\begin{figure}[t]
    \centering
    \includegraphics[width=1\linewidth]{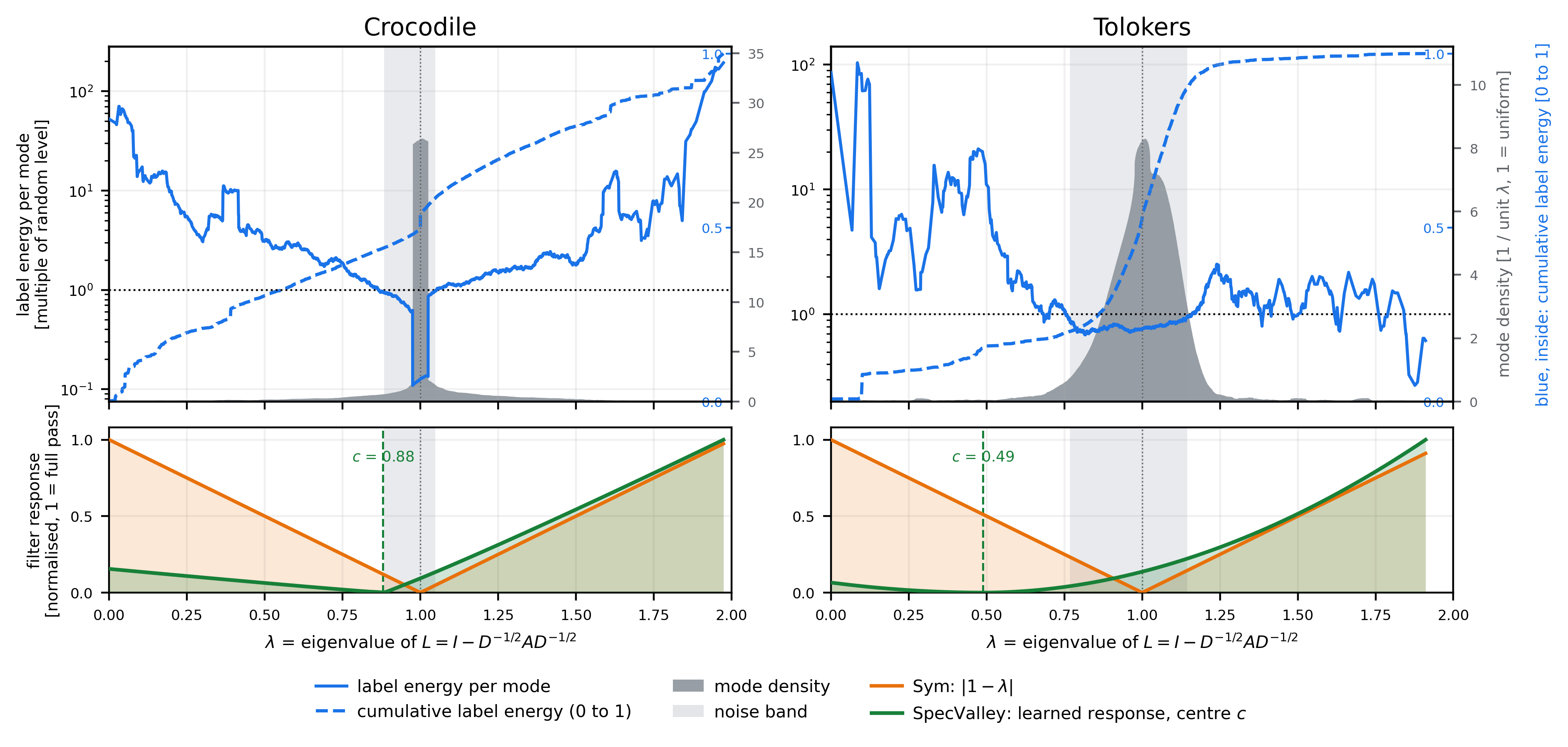}
    \vspace{-8mm}
    \caption{Label energy across the Laplacian spectrum (top) and the filter
    magnitude $|g(\lambda)|$ of degree normalisation and of SpecValley (bottom),
    for two dense graphs. Quantities are defined in the text; the remaining four
    graphs are in Appendix~\ref{app:spectral_aggregation_all_datasets}.}
    \label{fig:specvalley}
\end{figure}

To see where aggregation acts, we diagonalise
$L=I-D^{-1/2}AD^{-1/2}=U\Lambda U^{\top}$ and measure the share of label energy
each mode carries, $\hat e_k=\lVert u_k^{\top}Y\rVert_2^2/\lVert Y\rVert_F^2$
with $\sum_k \hat e_k=1$ (Appendix~\ref{app:spectral_aggregation_all_datasets}).
Figure~\ref{fig:specvalley} plots $n\,\hat e_k$, the multiple of the level a
random labelling would produce, together with the cumulative share, the
eigenvalue density and a band marking modes significantly below the random
level. The lower panel contrasts the fixed response $1-\lambda$ of degree
normalisation (Proposition~\ref{prop:dense_valley}) with the learned response
$r(\lambda)$ of Eq.~\eqref{eq:valley_response}, which recovers it for $c=p=1$ and
$a_L=-a_R=1$.

The two graphs have nearly the same spectrum: as
Proposition~\ref{prop:dense_valley} predicts for dense graphs, the eigenvalues
pile up at $\lambda=1$, with $76\%$ and $66\%$ of the modes in
$|\lambda-1|<0.1$. The label energy tells them apart. On Crocodile this core
carries only $17\%$ of the label energy, so attenuating it removes mostly noise,
and degree normalisation helps under the polynomial filters
(Table~\ref{tab:ablation_spectral}). On Tolokers it carries $52\%$, so
attenuation removes signal as well; there degree normalisation hurts under the
same filters, and the learned valley moves its centre away from $\lambda=1$.
Mode crowding alone therefore does not decide whether aggregation helps; the
label energy in the attenuated band matters. It is not a complete predictor,
however: on HM-categories, whose core also carries most of the label energy,
degree normalisation helps under every filter.

\subsection{Generalisation, Robustness and Filter Order}
\label{sec:sparsification}

\begin{figure}[t]
    \centering
    \includegraphics[width=1\linewidth]{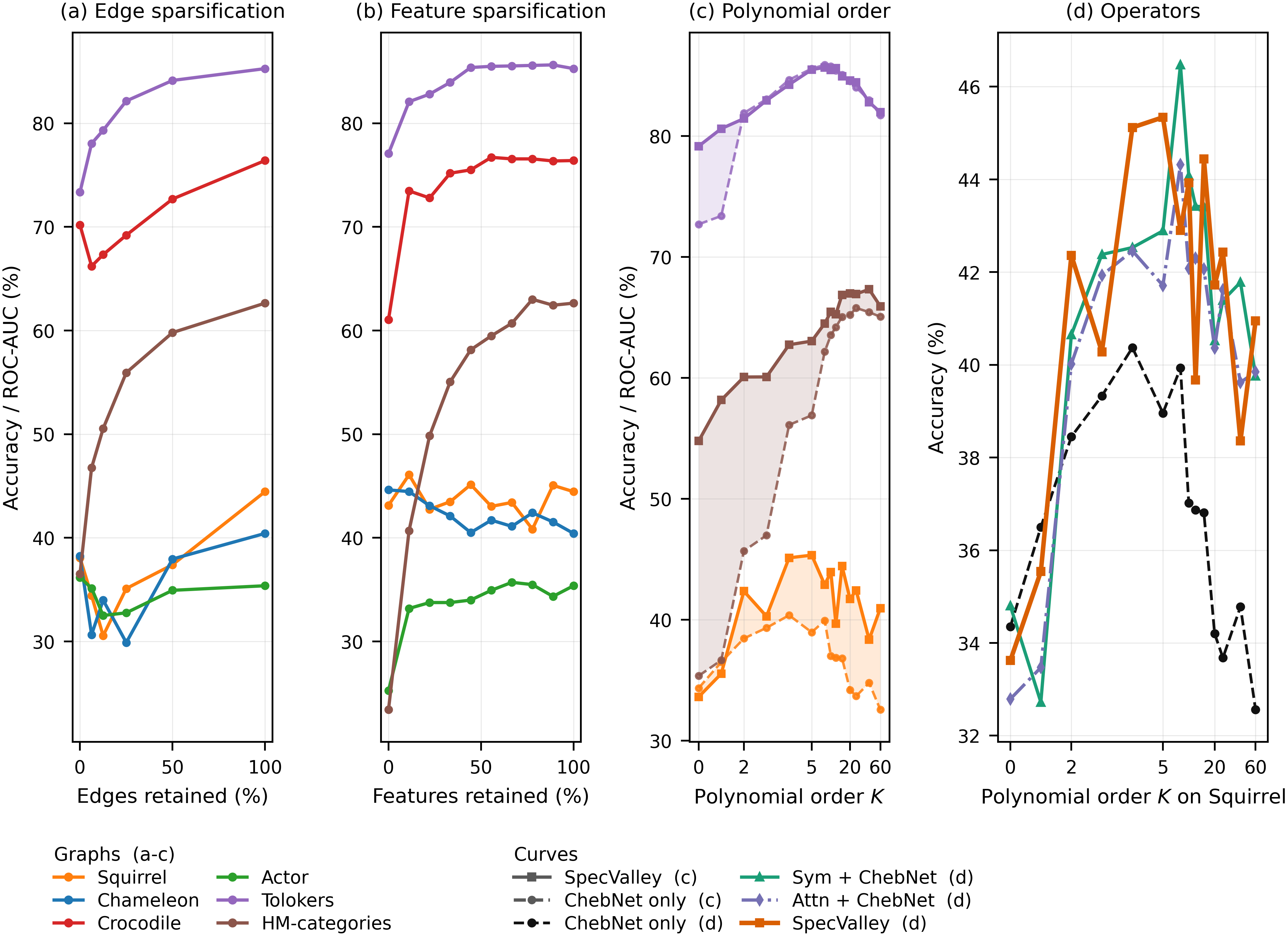}

    \vspace{-2mm}
    \caption{Robustness and filter order. (a,b) Edges and features removed
    progressively, SpecValley with ChebNet. (c) Polynomial order $K$ of the
    ChebNet layer with and without the valley aggregator, all six graphs.
    (d) The same sweep on Squirrel-f for four aggregation operators. Three
    splits (one on HM-categories), 150 epochs.}
    \label{fig:Sparsification_Plots}
\end{figure}

We probe SpecValley along three axes that leave the protocol of
Section~\ref{sec:experimental_settings} otherwise unchanged
(Figure~\ref{fig:Sparsification_Plots}): the graph is thinned by halving the
edges in six steps from $100\%$ to $0\%$; the features are reduced in ten equal
steps over the same range; and the order $K$ of the downstream ChebNet layer
(Eq.~\eqref{eq:cheb_conv}) is swept over $K \in \{0,1,2,3,4,5,8,10,12,15,20,25,40,60\}$.

\textbf{Sparsification.} Panel (a) shows that the effectiveness of SpecValley
rests on the density of the edges: every halving costs performance, but
the loss stays small down to about half the edges and grows only below that,
so the scores remain relatively stable over a wide range. Panel (b) shows
the converse: apart from HM-categories and from very strong reductions, the
scores stay almost constant across the whole range, so the label information
resides in the structure rather than in the feature dimension.

\textbf{Filter order.} Panel (c) shows that the filter does not absorb the
aggregation step. On the graphs where the valley is effective (Squirrel-f,
Chameleon-f, Crocodile and HM-categories) the gap to ChebNet alone does not
close with $K$, even at orders far beyond those used in the benchmark. The
composition of aggregator and filter is a polynomial of only one order higher
(Section~\ref{subsec:Compatibility_Aggregation_Filter}), so this is not a
question of expressive capacity but of what the two parameterisations reach in
training. On Tolokers and Actor the two curves coincide from moderate orders
onward, consistent with spectra in which no aggregator finds a band to attenuate
(Section~\ref{subsec:spectral_properties_datasets}). Panel (d) repeats the sweep on Squirrel-f with four
operators. All three aggregators stay above ChebNet alone across the whole
range, and the spread between them is small next to that gap, so at fixed $K$ it
is the presence of an aggregation step, more than its particular form, that
accounts for the difference. Beyond $K \approx 20$ every curve declines,
including ChebNet alone. Further results are reported in
Appendix~\ref{app:polynomial_order}.

\newpage
\section{Conclusion}
\label{sec:conclusion}
The aggregation step that precedes every graph filter is a spectral operator
in its own right, and on dense graphs its fixed zero falls exactly where the
eigenvalues accumulate. Making this step learnable is enough to change the
outcome: SpecValley, a learned valley in front of an unchanged ChebNet,
is the best
on four of six heterophilic benchmarks and within the confidence interval of the best on a fifth,
and the ablation over aggregators and filters
shows that no downstream filter closes the gap on its own. The label-energy
spectrum explains where the valley helps and where, as on Actor, no spectral
aggregator can.

\textbf{Limitations and Future Work.}
Our analysis is confined to dense heterophilic graphs with informative
spectral structure and to the linear part of the pipeline; ReLU and layer
normalisation can move energy between modes, and the operator parameters are
fitted on half of the validation set. Two questions remain open: when
attention before propagation aligns with the spectrum and when it learns
node-specific weights, and why a Chebyshev filter does not reproduce the gain
of the valley by itself, although the composition of valley and filter is a
polynomial of only one order higher.

\section*{AI use statement}

We used generative AI tools for research ideation, method development, for
writing experimental code, for writing and editing the text and for generating
the figures. We have reviewed the AI-assisted work and the final
content of this work.

\section*{Reproducibility Statement}
All datasets are public (Section~\ref{sec:experimental_settings}). The stored splits, the training protocol and all hyperparameters are given in Appendix~\ref{app:implementation_details} and complete proofs in Appendix~\ref{app:theoretical_analysis}. Code and splits are included with the submission.

\bibliography{GRIPedge}
\bibliographystyle{unsrtnat}
\newpage
\appendix
\onecolumn
\section{Theoretical Analysis}
\label{app:theoretical_analysis}

\subsection{Proof of Proposition~\ref{prop:dense_valley} (Uniform Aggregation as a Spectral Valley)}
\label{app:lemma_valley}
Let $A \in \mathbb{R}^{n\times n}$ be symmetric with nonnegative entries,
without self-loops ($A_{ii} = 0$) and without isolated nodes, i.e.\ with degrees
$d_i := \sum_j A_{ij} > 0$ for all $i$; let
$D := \mathrm{diag}(d_1,\dots,d_n)$, $d_{\min} := \min_i d_i$ and
$d_{\max} := \max_i d_i$. We call the graph $r$-regular if $d_i = r$ for all
$i$, writing $r$ for the common degree since $D$ denotes the degree matrix. Let
\[
P_{\mathrm{sym}} := D^{-1/2}AD^{-1/2},
\qquad
L := I - P_{\mathrm{sym}} = U\Lambda U^\top,
\qquad
d_{\mathrm{eff}} := \frac{n}{\|P_{\mathrm{sym}}\|_F^2},
\]
where $U = [u_1,\dots,u_n]$ is orthogonal and
$\Lambda = \mathrm{diag}(\lambda_1,\dots,\lambda_n)$ lists the eigenvalues of
$L$ with multiplicity (this decomposition exists by Step~0 below). We call $u_k$
the $k$-th mode, write $e_k$ for the $k$-th standard basis vector, and for
$q : \sigma(L) \to \mathbb{R}$ define the spectral function
$q(L) := U\,\mathrm{diag}\big(q(\lambda_1),\dots,q(\lambda_n)\big)\,U^\top$.

\paragraph{Proposition~\ref{prop:dense_valley}.}\label{prop:dense_valley-app}
In the setting above, $\frac{1}{n}\sum_k \lambda_k = 1$ and
$\frac{1}{n}\sum_k (\lambda_k - 1)^2 = 1/d_{\mathrm{eff}}$, and
$P_{\mathrm{sym}} = U(I-\Lambda)U^\top$ has the spectral response $1-\lambda$,
which vanishes at this mean. If moreover $d_{\mathrm{eff}} \ge 1/(\beta^2\eta)$
for some $\beta > 0$ and $\eta \in (0,1)$, then at least $(1-\eta)\,n$ of the
eigenvalues $\lambda_1,\dots,\lambda_n$ satisfy $|\lambda_k - 1| < \beta$, and
$P_{\mathrm{sym}}$ scales each of the corresponding modes by a factor of
absolute value less than $\beta$. For unweighted $A$,
$d_{\min} \le d_{\mathrm{eff}} \le d_{\max}$, with $d_{\mathrm{eff}} = r$ on
$r$-regular graphs.

\begin{proof}
Step~0 checks that all objects are well defined, Steps~1--3 establish the first
two moments of the spectrum and the response of $P_{\mathrm{sym}}$, Steps~4--5
the concentration and the attenuation, and Step~6 the degree bounds.

\medskip\noindent\textit{Step 0 (well-posedness).}
Since every $d_i > 0$, $D^{-1/2}$ exists. As $A$ is symmetric and $D$ diagonal,
$P_{\mathrm{sym}}^\top = D^{-1/2}A^\top D^{-1/2} = P_{\mathrm{sym}}$, so
$P_{\mathrm{sym}}$ and $L$ are real symmetric, and by the spectral theorem
$L = U\Lambda U^\top$ with orthogonal $U$ and real $\lambda_k$. Since
$d_1 = \sum_j A_{1j} > 0$ and $A_{1j} \ge 0$, some entry of $A$ is positive, so
$\|P_{\mathrm{sym}}\|_F^2 = \sum_{i,j} A_{ij}^2/(d_i d_j) > 0$ and
$d_{\mathrm{eff}} \in (0,\infty)$.

\medskip\noindent\textit{Step 1 (mean).}
Since $A_{ii} = 0$, the diagonal of $L$ is
$L_{ii} = 1 - (P_{\mathrm{sym}})_{ii} = 1 - A_{ii}/d_i = 1$, so
$\operatorname{tr}(L) = n$. By cyclicity of the trace and $U^\top U = I$,
$\operatorname{tr}(L) = \operatorname{tr}(U\Lambda U^\top)
= \operatorname{tr}(\Lambda U^\top U) = \sum_k \lambda_k$. Hence
$\frac{1}{n}\sum_k \lambda_k = 1$.

\medskip\noindent\textit{Step 2 (spectral form and mode-wise action).}
Since $U$ is square and orthogonal, $I = UU^\top$, and
\begin{equation}\label{eq:valley-psym}
P_{\mathrm{sym}} \;=\; I - L \;=\; UU^\top - U\Lambda U^\top
\;=\; U(I-\Lambda)U^\top ,
\end{equation}
so $P_{\mathrm{sym}} = q(L)$ with $q(\lambda) = 1-\lambda$. With
$U^\top u_k = e_k$, this gives $P_{\mathrm{sym}}u_k = U(I-\Lambda)e_k
= (1-\lambda_k)\,u_k$. By symmetry of $P_{\mathrm{sym}}$ (Step~0), for every
$Y \in \mathbb{R}^{n\times F}$,
\begin{equation}\label{eq:valley-mode}
u_k^\top P_{\mathrm{sym}}Y \;=\; (P_{\mathrm{sym}}u_k)^\top Y
\;=\; (1-\lambda_k)\,u_k^\top Y ,
\end{equation}
i.e.\ $P_{\mathrm{sym}}$ multiplies the component of every signal along $u_k$ by
$1-\lambda_k$. The response $1-\lambda$ vanishes if and only if $\lambda = 1$,
the mean of Step~1.

\medskip\noindent\textit{Step 3 (mean-square deviation).}
For every $M \in \mathbb{R}^{n\times F}$, orthogonality of $U$ gives
\begin{equation}\label{eq:valley-parseval}
\|M\|_F^2 \;=\; \operatorname{tr}(M^\top M)
\;=\; \operatorname{tr}(M^\top UU^\top M) \;=\; \|U^\top M\|_F^2
\;=\; \sum_{k=1}^n \|u_k^\top M\|_2^2 ,
\end{equation}
since the $k$-th row of $U^\top M$ is $u_k^\top M$; this is Parseval's identity
in the graph Fourier basis. For $M = P_{\mathrm{sym}}$,
\eqref{eq:valley-mode} with $Y = I$ gives
$u_k^\top P_{\mathrm{sym}} = (1-\lambda_k)\,u_k^\top$, hence
$\|u_k^\top P_{\mathrm{sym}}\|_2^2 = (1-\lambda_k)^2\|u_k\|_2^2
= (\lambda_k - 1)^2$, and \eqref{eq:valley-parseval} yields
\[
\|P_{\mathrm{sym}}\|_F^2 \;=\; \sum_{k=1}^n (\lambda_k - 1)^2 ,
\qquad\text{i.e.}\qquad
\frac{1}{n}\sum_k(\lambda_k-1)^2 \;=\; \frac{\|P_{\mathrm{sym}}\|_F^2}{n}
\;=\; \frac{1}{d_{\mathrm{eff}}} .
\]

\medskip\noindent\textit{Step 4 (concentration).}
Let $K_\beta := \{k : |\lambda_k - 1| \ge \beta\}$. Every $k \in K_\beta$
contributes at least $\beta^2$ to the sum of Step~3, and all other terms are
nonnegative, so
\[
\beta^2\,|K_\beta| \;\le\; \sum_{k \in K_\beta} (\lambda_k - 1)^2
\;\le\; \sum_{k=1}^n (\lambda_k - 1)^2 \;=\; \frac{n}{d_{\mathrm{eff}}} .
\]
This is Chebyshev's inequality for the uniform distribution over
$\lambda_1,\dots,\lambda_n$, whose mean is $1$ and whose variance is
$1/d_{\mathrm{eff}}$ by Steps~1 and~3. Since $\beta > 0$, $d_{\mathrm{eff}} > 0$
and $d_{\mathrm{eff}} \ge 1/(\beta^2\eta)$, it follows that
$|K_\beta| \le n/(\beta^2 d_{\mathrm{eff}}) \le \eta\,n$, so at least
$n - |K_\beta| \ge (1-\eta)\,n$ indices satisfy $|\lambda_k - 1| < \beta$.

\medskip\noindent\textit{Step 5 (attenuation).}
For every $k \notin K_\beta$, $|1-\lambda_k| = |\lambda_k - 1| < \beta$, so by
\eqref{eq:valley-mode} $P_{\mathrm{sym}}$ scales the corresponding mode by a
factor of absolute value less than $\beta$.

\medskip\noindent\textit{Step 6 (degree bounds).}
For unweighted $A$, $A_{ij} \in \{0,1\}$ gives $A_{ij}^2 = A_{ij}$, and with
$\mathcal{N}(i) := \{j : A_{ij} = 1\}$, so that $|\mathcal{N}(i)| = d_i$,
\[
\|P_{\mathrm{sym}}\|_F^2 \;=\; \sum_{i,j}\frac{A_{ij}}{d_i d_j}
\;=\; \sum_{i=1}^n \frac{1}{d_i} \sum_{j \in \mathcal{N}(i)} \frac{1}{d_j} .
\]
The inner sum has $d_i$ terms, each in $[1/d_{\max},\, 1/d_{\min}]$, so it lies
in $[d_i/d_{\max},\, d_i/d_{\min}]$, and each outer summand lies in
$[1/d_{\max},\, 1/d_{\min}]$. Summing over the $n$ nodes,
$n/d_{\max} \le \|P_{\mathrm{sym}}\|_F^2 \le n/d_{\min}$, i.e.\
$d_{\min} \le d_{\mathrm{eff}} \le d_{\max}$. On an $r$-regular graph every
outer summand equals $\frac{1}{r}\cdot r\cdot\frac{1}{r} = \frac{1}{r}$, hence
$\|P_{\mathrm{sym}}\|_F^2 = n/r$ and $d_{\mathrm{eff}} = r$.
\end{proof}

\paragraph{Remark (noise).}
As in Step~2, every spectral function satisfies
$u_k^\top q(L)\,Y = q(\lambda_k)\,u_k^\top Y$, so $q(L)$ scales the energy of
mode $k$ by $q(\lambda_k)^2$, for signal and noise alike. Let
$N \in \mathbb{R}^{n\times F}$ have i.i.d.\ zero-mean entries of variance
$\sigma^2 > 0$. Since $\mathbb{E}[N_{:j}N_{:j}^\top] = \sigma^2 I$ for every
column,
\begin{equation}\label{eq:valley-noise}
\mathbb{E}\|PN\|_F^2 \;=\; F\sigma^2\,\|P\|_F^2
\qquad\text{for every } P \in \mathbb{R}^{n\times n}.
\end{equation}
With $P = P_{\mathrm{sym}}$ and Step~3,
$\mathbb{E}\|P_{\mathrm{sym}}N\|_F^2 = \mathbb{E}\|N\|_F^2/d_{\mathrm{eff}}$:
uniform aggregation retains exactly a fraction $1/d_{\mathrm{eff}}$ of the
expected noise energy. With $P = u_k^\top$,
$\mathbb{E}\|u_k^\top N\|_2^2 = F\sigma^2$ for every $k$: isotropic noise
spreads evenly over the modes. If a signal $S$ spreads evenly as well,
$\|u_k^\top S\|_2^2 = \|S\|_F^2/n$ for all $k$, then by
\eqref{eq:valley-parseval} every $q(L)$ with $\sum_k q(\lambda_k)^2 > 0$ scales
the signal energy and the expected noise energy by the same factor
$\frac{1}{n}\sum_k q(\lambda_k)^2$, so no spectral aggregator changes their
ratio.

\paragraph{Remark (mean aggregation).}
$P_{\mathrm{mean}} := D^{-1}A = D^{-1/2}P_{\mathrm{sym}}D^{1/2}$ is similar to
$P_{\mathrm{sym}}$ and has the same eigenvalues $1-\lambda_k$, but it is
symmetric, and then equal to $P_{\mathrm{sym}}$, only if the degree is constant
on every connected component. Otherwise it is not a spectral function of $L$,
and the proposition applies to it only through its eigenvalues.

\paragraph{Remark (provenance).}
Steps~1--3 are standard facts about the normalised
Laplacian~\citep{chung1997spectral}, and Step~4 is Chebyshev's inequality. What
the proposition adds is the reading of uniform aggregation as a \emph{fixed
spectral valley} whose zero coincides with the mean of the spectrum, together
with the effective degree $d_{\mathrm{eff}}$ as the quantity that controls how
many modes lie in that valley. For random graphs with minimum expected degree
well above $\log^2 n$, all non-trivial eigenvalues concentrate at
$1$~\citep{chung2003spectra}; the proposition gives a weaker but deterministic
statement for every graph. The $1/d_{\mathrm{eff}}$ noise reduction is the
energy counterpart of the $1/\sqrt{\bar d}$ gain in separability in the expected
degree $\bar d$ of~\citet{baranwal2021graph}.

\paragraph{Remark (what the proposition does not claim).}
The bound in Step~4 uses only the second moment of the spectrum and is therefore
conservative; it is informative only for $d_{\mathrm{eff}} > 1/\beta^2$, since
otherwise $n/(\beta^2 d_{\mathrm{eff}}) \ge n$. The effective degree is not the
mean degree: on a star with $n \ge 3$ nodes the mean degree is
$2(n-1)/n < 2$, but $\|P_{\mathrm{sym}}\|_F^2 = 2$ and hence
$d_{\mathrm{eff}} = n/2$, and indeed $n-2$ eigenvalues equal $1$ exactly. The
proposition locates the modes, not the labels: whether attenuating the core
removes more noise than label information depends on the task, and no statement
about accuracy follows. The noise remark assumes isotropic noise; noise
correlated with the graph does not spread evenly over the modes.

\newpage
\subsection{Proof of Corollary~\ref{cor:energy_budget} (The Spectral Energy Budget of Aggregation)}
\label{app:proposition_energy}

Let $L = U\Lambda U^\top$ be as in Appendix~\ref{app:lemma_valley}, with
orthogonal $U = [u_1,\dots,u_n]$ and
$\Lambda = \mathrm{diag}(\lambda_1,\dots,\lambda_n)$. Fix \emph{a priori} a
partition $\{\mathcal{I}_b\}_{b=1}^m$ of $\{1,\dots,n\}$ into nonempty bands, and
set
\[
P_b := \sum_{k \in \mathcal{I}_b} u_k u_k^\top,
\qquad
E_b(Y) := \|P_b Y\|_F^2
\quad\text{for } Y \in \mathbb{R}^{n\times F}.
\]
Let $\mathcal{G}_\Gamma := \{\, g : \sigma(L) \to \mathbb{R} \,:\,
\max_k |g(\lambda_k)| \le \Gamma \,\}$, and for $g \in \mathcal{G}_\Gamma$ put
$g(L) := U g(\Lambda) U^\top$ with
$g(\Lambda) := \mathrm{diag}\big(g(\lambda_1),\dots,g(\lambda_n)\big)$,
\[
\mu_b^-(g) := \min_{k \in \mathcal{I}_b} |g(\lambda_k)|,
\qquad
\mu_b^+(g) := \max_{k \in \mathcal{I}_b} |g(\lambda_k)| ;
\]
$g$ is \emph{band-constant} if $\mu_b^-(g) = \mu_b^+(g) =: \mu_b(g)$ for every
$b$.

\paragraph{Corollary~\ref{cor:energy_budget}.}\label{cor:energy-app}
In the setting above, let $Y \in \mathbb{R}^{n\times F}$ be a graph signal, such
as the aggregated features $Y = A(x)X$, and $Z := g(L)\,Y$ with
$g \in \mathcal{G}_\Gamma$. Then:

\emph{(i) Band-wise budget.}
$\|P_b Z\|_F^2 = \sum_{k \in \mathcal{I}_b} g(\lambda_k)^2 \|u_k^\top Y\|_2^2$,
hence $\big(\mu_b^-(g)\big)^2 E_b(Y) \le \|P_b Z\|_F^2 \le
\big(\mu_b^+(g)\big)^2 E_b(Y)$, with equality for band-constant $g$.

\emph{(ii) Non-creation.} If $E_b(Y) = 0$, then $P_b Z = 0$ for every
$g \in \mathcal{G}_\Gamma$.

\begin{proof}
For an index set $\mathcal{I} \subseteq \{1,\dots,n\}$ write
$\Pi_{\mathcal{I}} := \sum_{k \in \mathcal{I}} e_k e_k^\top$ for its coordinate
selector, and $\Pi_b := \Pi_{\mathcal{I}_b}$. Steps~0--3 collect the properties
of the projections, Steps~4--6 prove (i) and (ii), and Steps~7--8 prove the two
statements of the main text that follow the corollary: the central-band bound
and the absence of cross-band transfer for spectral aggregators.

\medskip\noindent\textit{Step 0 (well-posedness).}
$U$ is orthogonal by Step~0 of Appendix~\ref{app:lemma_valley}. Every
$g \in \mathcal{G}_\Gamma$ is defined on $\sigma(L)$, so $g(\Lambda)$ and $g(L)$
are well defined, and since every band is nonempty, $\mu_b^\pm(g)$ are minima and
maxima over nonempty finite sets.

\medskip\noindent\textit{Step 1 (projections in the eigenbasis).}
Since $u_k = Ue_k$,
\begin{equation}\label{eq:proj-eigenbasis}
P_b \;=\; \sum_{k \in \mathcal{I}_b} U e_k e_k^\top U^\top \;=\; U \Pi_b U^\top .
\end{equation}
As $\{\mathcal{I}_b\}$ partitions $\{1,\dots,n\}$, $\sum_b \Pi_b = I$ and
$\Pi_b \Pi_{b'} = \delta_{bb'}\Pi_b$. With $U^\top U = UU^\top = I$ this gives
$\sum_b P_b = I$, $P_b P_{b'} = \delta_{bb'} P_b$, and $P_b^\top = P_b$.

\medskip\noindent\textit{Step 2 (commutation with spectral functions).}
Let $h : \sigma(L) \to \mathbb{R}$ and $h(L) := U h(\Lambda) U^\top$. By
\eqref{eq:proj-eigenbasis} and $U^\top U = I$,
$P_b\, h(L) = U \Pi_b h(\Lambda) U^\top$ and
$h(L)\, P_b = U h(\Lambda) \Pi_b U^\top$. As $\Pi_b$ and $h(\Lambda)$ are
diagonal, they commute, hence
\begin{equation}\label{eq:commute}
P_b\, h(L) \;=\; h(L)\, P_b \qquad \text{for every } b \text{ and every } h .
\end{equation}

\medskip\noindent\textit{Step 3 (band energy in the modes).}
For every $M \in \mathbb{R}^{n\times F}$, \eqref{eq:proj-eigenbasis} gives
$U^\top P_b M = \Pi_b U^\top M$, whose $k$-th row is $u_k^\top M$ for
$k \in \mathcal{I}_b$ and zero otherwise. With the Frobenius isometry
$\|P_b M\|_F = \|U^\top P_b M\|_F$ of \eqref{eq:valley-parseval},
\begin{equation}\label{eq:band-energy-def}
E_b(M) \;=\; \|P_b M\|_F^2 \;=\; \sum_{k \in \mathcal{I}_b} \|u_k^\top M\|_2^2 .
\end{equation}

\medskip\noindent\textit{Step 4 (identity of part (i)).}
By \eqref{eq:proj-eigenbasis}, $U^\top U = I$ and $g(L) = Ug(\Lambda)U^\top$,
\[
U^\top P_b Z \;=\; U^\top\,U\Pi_bU^\top\,U g(\Lambda)U^\top\,Y
\;=\; \Pi_b\, g(\Lambda)\, U^\top Y ,
\]
whose $k$-th row is $g(\lambda_k)\,u_k^\top Y$ for $k \in \mathcal{I}_b$ and
zero otherwise. Since $\|P_b Z\|_F = \|U^\top P_b Z\|_F$ and each
$g(\lambda_k)$ is a scalar,
\begin{equation}\label{eq:filtered-band}
\|P_b Z\|_F^2 \;=\; \sum_{k \in \mathcal{I}_b}
\big\|g(\lambda_k)\,u_k^\top Y\big\|_2^2
\;=\; \sum_{k \in \mathcal{I}_b} g(\lambda_k)^2\,\|u_k^\top Y\|_2^2 .
\end{equation}
Steps~3 and~4 use neither the bound $\Gamma$ nor the partition property: they
hold verbatim for every function $g$ on $\sigma(L)$ and for every index set
$\mathcal{I}$ in place of $\mathcal{I}_b$, with
$P_{\mathcal{I}} := U\Pi_{\mathcal{I}}U^\top$ and
$E_{\mathcal{I}}(M) := \|P_{\mathcal{I}}M\|_F^2$.

\medskip\noindent\textit{Step 5 (two-sided bound and equality case).}
For $k \in \mathcal{I}_b$, the definition of $\mu_b^\pm(g)$ gives
$\mu_b^-(g) \le |g(\lambda_k)| \le \mu_b^+(g)$, and squaring preserves these
inequalities between nonnegative numbers. Bounding every summand of
\eqref{eq:filtered-band} on both sides and using
$\sum_{k \in \mathcal{I}_b}\|u_k^\top Y\|_2^2 = E_b(Y)$ from
\eqref{eq:band-energy-def},
\[
\big(\mu_b^-(g)\big)^2\, E_b(Y) \;\le\; \|P_b Z\|_F^2
\;\le\; \big(\mu_b^+(g)\big)^2\, E_b(Y) .
\]
If $g$ is band-constant, both bounds equal $\mu_b(g)^2 E_b(Y)$, so
$\|P_b Z\|_F^2 = \mu_b(g)^2 E_b(Y)$. This proves (i).

\medskip\noindent\textit{Step 6 (part (ii)).}
If $E_b(Y) = 0$, the upper bound of Step~5 gives $\|P_b Z\|_F^2 \le 0$, hence
$P_b Z = 0$. Only the finiteness of $\mu_b^+(g)$ was used, so the conclusion
holds for every $g \in \mathcal{G}_\Gamma$ simultaneously. This proves (ii).

\medskip\noindent\textit{Step 7 (central band).}
Let $\beta > 0$, $\mathcal{I}_\beta := \{k : |\lambda_k - 1| < \beta\}$,
$P_\beta := P_{\mathcal{I}_\beta}$ and $E_\beta := E_{\mathcal{I}_\beta}$. For
$Y = P_{\mathrm{sym}}X$, \eqref{eq:valley-mode} gives
$u_k^\top Y = (1-\lambda_k)\,u_k^\top X$. By \eqref{eq:filtered-band} for the
index set $\mathcal{I}_\beta$ (Step~4), $g(\lambda_k)^2 \le \Gamma^2$, and
$(1-\lambda_k)^2 < \beta^2$ for $k \in \mathcal{I}_\beta$,
\[
\|P_\beta Z\|_F^2 \;=\; \sum_{k \in \mathcal{I}_\beta}
g(\lambda_k)^2 (1-\lambda_k)^2 \|u_k^\top X\|_2^2
\;\le\; \Gamma^2\beta^2 \sum_{k \in \mathcal{I}_\beta}\|u_k^\top X\|_2^2
\;=\; \Gamma^2\beta^2\, E_\beta(X)
\]
for every $g \in \mathcal{G}_\Gamma$, using \eqref{eq:band-energy-def} for
$\mathcal{I}_\beta$ in the last step. If $\mathcal{I}_\beta = \emptyset$, both
sides are zero.

\medskip\noindent\textit{Step 8 (cross-band transfer).}
Let $q : \sigma(L) \to \mathbb{R}$. For $b \neq b'$, \eqref{eq:commute} with
$h = q$ and Step~1 give $P_b\, q(L)\, P_{b'} = q(L)\, P_b P_{b'} = 0$: a
spectral aggregator maps no energy from band $b'$ into band $b$, and
$P_b\, q(L)X = q(L)\,P_b X$ depends on $X$ only through $P_b X$. Since Step~4
does not use $\Gamma$, it applies with $g$ replaced by $q$ and $Y = X$, giving
$E_b\big(q(L)X\big) = \sum_{k \in \mathcal{I}_b} q(\lambda_k)^2\,
\|u_k^\top X\|_2^2$. For a general $A(x) \in \mathbb{R}^{n\times n}$, the cross-band blocks
$P_b A(x) P_{b'}$, $b \neq b'$, need not vanish, and energy can move between bands.
\end{proof}

\paragraph{Remark (provenance).}
Parts (i) and (ii) are elementary consequences of Parseval's identity and of
\eqref{eq:commute}; we state them as a corollary because Step~7 combines them
with Proposition~\ref{prop:dense_valley}. Their use lies in the quantifier:
(ii) holds for \emph{every} admissible filter, which makes it an impossibility
statement about the composition of aggregation and filter.

\paragraph{Remark (multi-branch aggregators).}
Since (i) and (ii) hold for every signal $Y$, they apply to the output
$H' = XW_I + SXW_S + S^\top XW_T + R(L)XW_R$ of Eq.~\eqref{eq:valley_operator},
with the scalar gates absorbed into the projections. On undirected graphs
$S = S^\top = P_{\mathrm{sym}}$, and $R(L) = U r(\Lambda) U^\top$ is a spectral
function of $L$, as is its Chebyshev approximation, which is a polynomial in
$L$. Every branch operator is therefore spectral, and Step~8 gives
$P_b H' = (P_bX)W_I + P_{\mathrm{sym}}(P_bX)(W_S + W_T) + R(L)(P_bX)W_R$: the
band-$b$ component of $H'$ depends on $X$ only through $P_bX$, so the adaptive
valley moves no energy between bands. Its ego branch keeps every band of $X$
available to the filter: if $E_b(X) > 0$, then $P_bX \neq 0$, so
$P_bXv \neq 0$ for some $v \in \mathbb{R}^d$, and with $W_I := v\,e_1^\top$ and
$W_S = W_T = W_R = 0$ one obtains $E_b(H') = \|P_bXv\|_2^2 > 0$. Hence the
architecture never forces $E_b(H') = 0$ on a band where $E_b(X) > 0$, whereas
$P_{\mathrm{sym}}$ alone does so on every band consisting only of eigenvalues
equal to $1$.

\paragraph{Remark (scope).}
The corollary bounds the energy that reaches each band, not accuracy: energy in
a band may be present without being linearly decodable. The partition
$\{\mathcal{I}_b\}$ must be fixed before the band energies are inspected; chosen
post hoc, the statement is void. The equality case of (i) is an idealisation:
a ChebNet response of degree at most $K-1$ has constant magnitude on a band with
more than $2(K-1)$ distinct eigenvalues only if it is constant, since
$p^2 - \mu^2$ would have more roots than its degree; Step~5 bounds the
deviation.

\newpage
\section{GCN Layer Descriptions}
\label{app:GCN_Layer_Descriptions}

In this part of the Appendix, we present the linear spectral downstream GCN layers that follow the aggregation stage introduced in Section~\ref{sec:method} and are evaluated in Section~\ref{sec:ablations}. We adopt simplified implementations of the layers rather than detailed one-to-one full reproductions. This choice is made to avoid unnecessary architectural complexity and to emphasize the essential structural principles of the respective models. Consequently, not all implementations correspond exactly to their original paper versions in every technical detail.

\vspace{0.5cm}
\subsection{APPNP (Gasteiger, Bojchevski \& Günnemann, 2019)}
\label{app:APPNP}

\textbf{Mathematical Description:}

APPNP combines a learnable feature transformation with a personalized
PageRank-type diffusion process. First, an initial representation is computed as

\begin{equation}
\mathbf{H}^{(0)} = \mathbf{X} \mathbf{W},
\end{equation}

where $\mathbf{W}$ is a learnable weight matrix.
Propagation is then performed iteratively as

\begin{equation}
\mathbf{H}^{(k+1)}
=
(1-\alpha)\mathbf{H}^{(0)}
+
\alpha \mathbf{\tilde{A}} \mathbf{H}^{(k)},
\qquad k = 0, \dots, K-1,
\end{equation}

with final output

\begin{equation}
\mathbf{H}^{(out)} = \mathrm{ReLU}(\mathbf{H}^{(K)}).
\end{equation}

Here, $\mathbf{\tilde{A}}$ denotes the symmetrically normalized adjacency matrix

\begin{equation}
\mathbf{\tilde{A}} =
\mathbf{D}^{-1/2}
\mathbf{A}
\mathbf{D}^{-1/2}.
\end{equation}

This corresponds to a truncated personalized PageRank (PPR) diffusion.
In a PageRank-style process, each node's representation is updated as a weighted combination of its own features and the features of its neighbors. The parameter $\alpha$ controls the relative weight of the propagated neighbor information versus the node's original features, analogous to the teleport probability in classical PageRank. \cite{gasteiger_predict_2019}

\textbf{Legend:}
\begin{itemize}
    \item $\mathbf{X} \in \mathbb{R}^{N \times d}$: Input node feature matrix
    \item $\mathbf{W} \in \mathbb{R}^{d \times d'}$: Learnable linear transformation
    \item $\mathbf{H}^{(k)} \in \mathbb{R}^{N \times d'}$: Representation after $k$ propagation steps
    \item $\mathbf{A} \in \mathbb{R}^{N \times N}$: Adjacency matrix
    \item $\mathbf{D}$: Diagonal degree matrix
    \item $\mathbf{\tilde{A}}$: Symmetrically normalized adjacency matrix
    \item $\alpha \in (0,1)$: Teleport (diffusion) parameter, controlling the weight of the initial node features
    \item $K$: Number of propagation steps
\end{itemize}

\vspace{0.5cm}
\subsection{ARMA (Bianchi, Grattarola, Livi \& Alippi, 2022)}
\label{app:ARMA}

\textbf{Mathematical Description:}

The ARMA spectral layer performs a frequency-selective, edge-aware graph convolution by iteratively combining Laplacian-propagated features with transformed input features. First, a linear feature transformation is applied \cite{bianchi2022arma}:

\begin{equation}
\mathbf{H}^{(0)} = \mathbf{X} \mathbf{W},
\end{equation}

where $\mathbf{W}$ is a learnable weight matrix.
Propagation is then performed iteratively over $K$ steps:

\begin{equation}
\mathbf{H}^{(k+1)} = \alpha \mathbf{L} \mathbf{H}^{(k)} + (1 - \alpha) \mathbf{H}^{(0)},
\qquad k = 0, \dots, K-1,
\end{equation}

with final output

\begin{equation}
\mathbf{H}^{(out)} = \mathrm{ReLU}(\mathbf{H}^{(K)}).
\end{equation}

Here, $\mathbf{L}$ denotes the normalized Laplacian matrix of the graph.

\textbf{Legend:}
\begin{itemize}
    \item $\mathbf{X} \in \mathbb{R}^{N \times d}$: Input node feature matrix
    \item $\mathbf{W} \in \mathbb{R}^{d \times d'}$: Learnable linear transformation
    \item $\mathbf{H}^{(k)} \in \mathbb{R}^{N \times d'}$: Representation after $k$ propagation steps
    \item $\mathbf{L} \in \mathbb{R}^{N \times N}$: Normalized Laplacian matrix
    \item $\alpha \in (0,1)$: Weighting between Laplacian propagation and original features
    \item $K$: Number of propagation iterations
    \item ReLU$(\cdot)$: Rectified Linear Unit activation function
\end{itemize}

\vspace{0.5cm}
\subsection{Gegenbauer Spectral Layer}
\label{app:Gegenbauer}

\textbf{Mathematical Description:}

Our Gegenbauer layer constructs a spectral graph filter based on
Gegenbauer polynomials of the graph Laplacian. It implements a
polynomial spectral filter, where each polynomial order is associated
with learnable feature mixing, allowing flexible aggregation across orders.
This design is inspired by recent work on Gegenbauer-based spectral GNNs
and GegenConv operators, which generalize Chebyshev filters using orthogonal
Gegenbauer polynomials \cite{castro2024gegenbauer, goksu2025gegenbauernet}.
Let $\mathbf{L} \in \mathbb{R}^{N \times N}$ denote the (normalized) graph Laplacian
and $\mathbf{H} \in \mathbb{R}^{N \times d}$ the input features.

The filter is defined as a truncated Gegenbauer polynomial expansion

\begin{equation}
g(\mathbf{L}) \mathbf{H}
=
\sum_{k=0}^{K-1}
\mathbf{C}_k^{(\alpha)}(\mathbf{L}) \mathbf{H} \mathbf{W}_k,
\end{equation}

where $\mathbf{W}_k \in \mathbb{R}^{d \times d}$ are learnable weight matrices
and $\mathbf{C}_k^{(\alpha)}(\cdot)$ denotes the $k$-th order Gegenbauer polynomial
with parameter $\alpha > 0$.

The polynomials are defined recursively as

\begin{align}
\mathbf{C}_0^{(\alpha)}(\mathbf{L}) \mathbf{H}
&= \mathbf{H}, \\
\mathbf{C}_1^{(\alpha)}(\mathbf{L}) \mathbf{H}
&= -2\alpha \mathbf{L} \mathbf{H}, \\
\mathbf{C}_k^{(\alpha)}(\mathbf{L}) \mathbf{H}
&=
\frac{2(k+\alpha-1)}{k}
\mathbf{L}\mathbf{C}_{k-1}^{(\alpha)}(\mathbf{L})\mathbf{H}
-
\frac{k+2\alpha-2}{k}
\mathbf{C}_{k-2}^{(\alpha)}(\mathbf{L})\mathbf{H},
\end{align}
where the sign on the $\mathbf{C}_1$ term reflects the implementation's
use of $-\mathbf{L}$ (equivalently, $\hat{\mathbf{A}} - \mathbf{I}$) as
the polynomial argument so that the first-order term combines with
$\mathbf{C}_0$ as $(\mathbf{I} - 2\alpha\mathbf{L})\mathbf{H}$, which
recovers the propagation operator $\hat{\mathbf{A}}\mathbf{H}$ in the
restricted form below.

The final layer output is then
\begin{equation}
\mathbf{H}^{(out)}
=
\mathrm{ReLU}
\left(
\sum_{k=0}^{K-1}
\mathbf{C}_k^{(\alpha)}(\mathbf{L}) \mathbf{H} \mathbf{W}_k
\right).
\end{equation}
Thus, the model learns a polynomial spectral filter whose shape
is controlled by the order $K$, the learnable matrices $\mathbf{W}_k$,
and the Gegenbauer parameter $\alpha$.
Different choices of $\alpha$ interpolate between known polynomial bases
(e.g., Chebyshev-type cases).

\textbf{Legend:}
\begin{itemize}
    \item $\mathbf{H} \in \mathbb{R}^{N \times d}$: Input node feature matrix
    \item $\mathbf{L} \in \mathbb{R}^{N \times N}$: Graph Laplacian
    \item $\mathbf{W}_k \in \mathbb{R}^{d \times d}$: Learnable weight matrix for order $k$
    \item $\alpha > 0$: Gegenbauer parameter controlling spectral shape
    \item $K$: Polynomial order (number of spectral components)
\end{itemize}

\newpage
\section{ Experimental Results}
\subsection{Implementation Details}
\label{app:implementation_details}
We provide additional implementation details as a complement to the Experimental Settings section in the main paper ~\ref{sec:experimental_settings}.

\subsection*{Training Protocol}

All models are trained from scratch on each split, with hidden dimension $64$,
input dimension equal to the number of node features and output dimension equal
to the number of classes. We use Adam (learning rate $3\times10^{-3}$, weight
decay $5\times10^{-4}$) on the cross-entropy loss, with dropout $0.5$ in the
aggregation stage and in every downstream layer; on Tolokers, the only binary
task, the loss is class-weighted by the inverse class ratio. Nodes without a
label are excluded from all masks. All six graphs are treated as undirected;
where the source provides directed edges, we symmetrise them before computing
$L$, so that $S = S^\top = P_{\mathrm{sym}}$ in Eq.~\eqref{eq:valley_operator}.
We evaluate after every epoch and report the test score of the epoch with the best validation score; the test labels enter no
selection at any point. Table~\ref{tab:protocol} summarises the protocol of the
three measurements.

\begin{table}[h]
\centering
\small
\setlength{\tabcolsep}{4pt}
\begin{tabular}{llrrl}
\toprule
Dataset & Split & Repeats & Metric & Selection \\
\midrule
Squirrel-f     & 48/32/20, random   & 10 & Accuracy & validation epoch \\
Chameleon-f    & 48/32/20, random   & 10 & Accuracy & validation epoch \\
Crocodile      & 48/32/20, random   & 10 & Accuracy & validation epoch \\
Actor          & 48/32/20, random   & 10 & Accuracy & validation epoch \\
Tolokers       & 50/25/25, official & 10 & ROC-AUC  & validation epoch \\
HM-categories  & 10/10/80, official & 1 split, 3 seeds & Accuracy & validation epoch \\
\midrule
\multicolumn{5}{l}{\textit{Measurements}} \\
Section~\ref{sec:ablations} (ablation)        & 150 epochs  & \multicolumn{3}{l}{$K=5$ on every graph, all repeats above} \\
Section~\ref{sec:benchmark_results} (benchmark) & 1000 epochs & \multicolumn{3}{l}{$K$ per dataset (Table~\ref{tab:k_choice}), all repeats above} \\
Section~\ref{sec:sparsification} (robustness, order $K$) & 150 epochs & \multicolumn{3}{l}{$K=5$ or swept; 3 repeats, 1 on HM-categories} \\
\bottomrule
\end{tabular}
\caption{Evaluation protocol. Splits are stored once and reused across all
models and measurements. The epoch is always selected on the validation set.}
\label{tab:protocol}
\end{table}

For SpecValley the validation set is split in half once per run: the operator
parameters $\theta = (c, p, a_L, a_R, \tau)$ are trained on half $A$ by a second
Adam without weight decay, while the epoch is selected on half $B$, which never
enters the spectral fit. All other parameters follow the training loss. The
spectral branch $R(L)$ uses a cached eigendecomposition of $L$ for graphs with
at most $14{,}000$ nodes and a Chebyshev expansion of degree $40$ in $L-I$
above, which applies to HM-categories only. All experiments run on a single
NVIDIA RTX 3060 (12\,GB) with 32\,GB system RAM.

\begin{table}[h]
\centering
\small
\setlength{\tabcolsep}{4.5pt}
\begin{tabular}{lrrrrrr}
\toprule
 & Squirrel-f & Chameleon-f & Crocodile & Actor & Tolokers & HM-cat. \\
\midrule
\multicolumn{7}{l}{\textit{ChebNet}} \\
Selected $K$       & 4     & 2     & 40    & 2     & 8     & 25    \\
Score              & 40.37 & 40.05 & 75.72 & 36.86 & 85.88 & 65.78 \\
Score at $K{=}5$   & 38.96 & 35.63 & 72.56 & 35.66 & 85.57 & 56.90 \\
$\Delta$           & +1.41 & +4.42 & +3.16 & +1.20 & +0.31 & +8.88 \\
\midrule
\multicolumn{7}{l}{\textit{SpecValley}} \\
Selected $K$       & 5     & 20    & 25    & 4     & 8     & 40    \\
Score              & 45.34 & 43.42 & 77.13 & 36.16 & 85.67 & 67.32 \\
Score at $K{=}5$   & 45.34 & 42.56 & 76.29 & 35.92 & 85.49 & 63.04 \\
$\Delta$           & 0.00  & +0.86 & +0.84 & +0.24 & +0.18 & +4.28 \\
\bottomrule
\end{tabular}
\caption{Polynomial order $K$ of the ChebNet layer in the benchmark, for
ChebNet alone and for SpecValley, from a 150-epoch sweep. Scores are test
accuracy (ROC-AUC on Tolokers) at $K$ selected on the validation score
and at $K=5$, taken at the
epoch with the best validation score and averaged over the sweep splits;
$\Delta$ is the difference between the two.}
\label{tab:k_choice}
\end{table}

\subsection*{Choice of the Polynomial Order $K$ for the Benchmark}

For the benchmark in Section~\ref{sec:benchmark_results}, we set the order $K$ of
the downstream ChebNet layer per dataset, since performance varies strongly
with $K$ across graphs (Section~\ref{sec:sparsification},
Figure~\ref{fig:Sparsification_Plots}).
$K$ is selected in a preliminary sweep over $K \in \{0,1,2,3,4,5,8,10,12,15,20,25,40,60\}$, training ChebNet alone and SpecValley with ChebNet for 150 epochs per order on three splits (one on HM-categories). $K=0$ applies no filter and $K=1$ does not use the graph. For each model and graph we take the order with the highest mean validation score. The test scores in Table~\ref{tab:k_choice} are reported for information only and enter no selection.

\subsection*{Runtime}

Table~\ref{tab:runtime} reports the cost of SpecValley under the protocol of
Appendix~\ref{app:implementation_details}, once with $K=5$ on every graph as in
the ablation of Section~\ref{sec:ablations}, and once with the per-dataset order
of Table~\ref{tab:k_choice} as in the benchmark of
Section~\ref{sec:benchmark_results}. The eigendecomposition of $L$ is computed
once per graph and cached; it is not part of the per-epoch times.

Two regimes are visible, and they follow the complexity of
Section~\ref{subsec:complexity}. On the five graphs with at most $14{,}000$
nodes the spectral branch runs on the cached eigenbasis at $O(n^2 d')$ per pass,
and an epoch costs a fraction of a second. HM-categories falls on the other side
of the threshold and uses the degree-$40$ Chebyshev expansion at $O(Msd')$,
where its $21.5$ million edges dominate: one epoch takes $15.8$ seconds, about
$90$ times Tolokers at four times the node count. Raising $K$ is comparatively
cheap, because the input projection and the aggregation stage do not depend on
it: five times the order costs $2.3\times$ the epoch on Crocodile and
$1.7\times$ on HM-categories. A full split takes
41 minutes for one 150-epoch ablation split and $7.7$ hours for one
$1000$-epoch benchmark split, of which HM-categories accounts for more than $95\%$.

\begin{table}[h]
\centering
\small
\setlength{\tabcolsep}{5pt}
\begin{tabular}{lrrrrr}
\toprule
Dataset & Nodes & $K$ & Time/epoch (s) & Split @ 150 ep. & Split @ 1000 ep. \\
\midrule
\multicolumn{6}{l}{\textit{Ablation setting: $K=5$ on every graph}} \\
Chameleon-f   &    890 &  5 &  0.073 & 11.1\,s   & 73.4\,s \\
Squirrel-f    &  2{,}223 &  5 &  0.084 & 12.6\,s   & 84.1\,s \\
Actor         &  7{,}600 &  5 &  0.073 & 10.9\,s   & 72.7\,s \\
Crocodile     & 11{,}631 &  5 &  0.216 & 32.6\,s   &  3.6\,min \\
Tolokers      & 11{,}758 &  5 &  0.176 & 26.4\,s   &  2.9\,min \\
HM-categories & 46{,}563 &  5 & 15.76  & 39.4\,min &  4.4\,h \\
\cmidrule(l){4-6}
All six, one split & & &      & 41.0\,min &  4.6\,h \\
\midrule
\multicolumn{6}{l}{\textit{Benchmark setting: $K$ per dataset}} \\
Chameleon-f   &    890 & 20 &  0.199 & 29.9\,s   &  3.3\,min \\
Squirrel-f    &  2{,}223 &  5 &  0.074 & 11.2\,s   & 74.3\,s \\
Actor         &  7{,}600 &  4 &  0.064 &  9.7\,s   & 64.3\,s \\
Crocodile     & 11{,}631 & 25 &  0.497 & 74.7\,s   &  8.3\,min \\
Tolokers      & 11{,}758 &  8 &  0.240 & 36.0\,s   &  4.0\,min \\
HM-categories & 46{,}563 & 40 & 26.53  & 66.3\,min &  7.4\,h \\
\cmidrule(l){4-6}
All six, one split & & &      & 69.0\,min &  7.7\,h \\
\bottomrule
\end{tabular}
\caption{Training time of SpecValley on a single NVIDIA RTX 3060 (12\,GB). The
ablation trains for 150 epochs with $K=5$, the benchmark for 1000 epochs with
the per-dataset order of Table~\ref{tab:k_choice}; the eigendecomposition is
computed once per graph and excluded.}
\label{tab:runtime}
\end{table}

\clearpage
\subsection{Spectral Valley Aggregation for all Datasets}
\label{app:spectral_aggregation_all_datasets}

\begin{figure}[ht]
    \centering
    \includegraphics[width=1\linewidth]{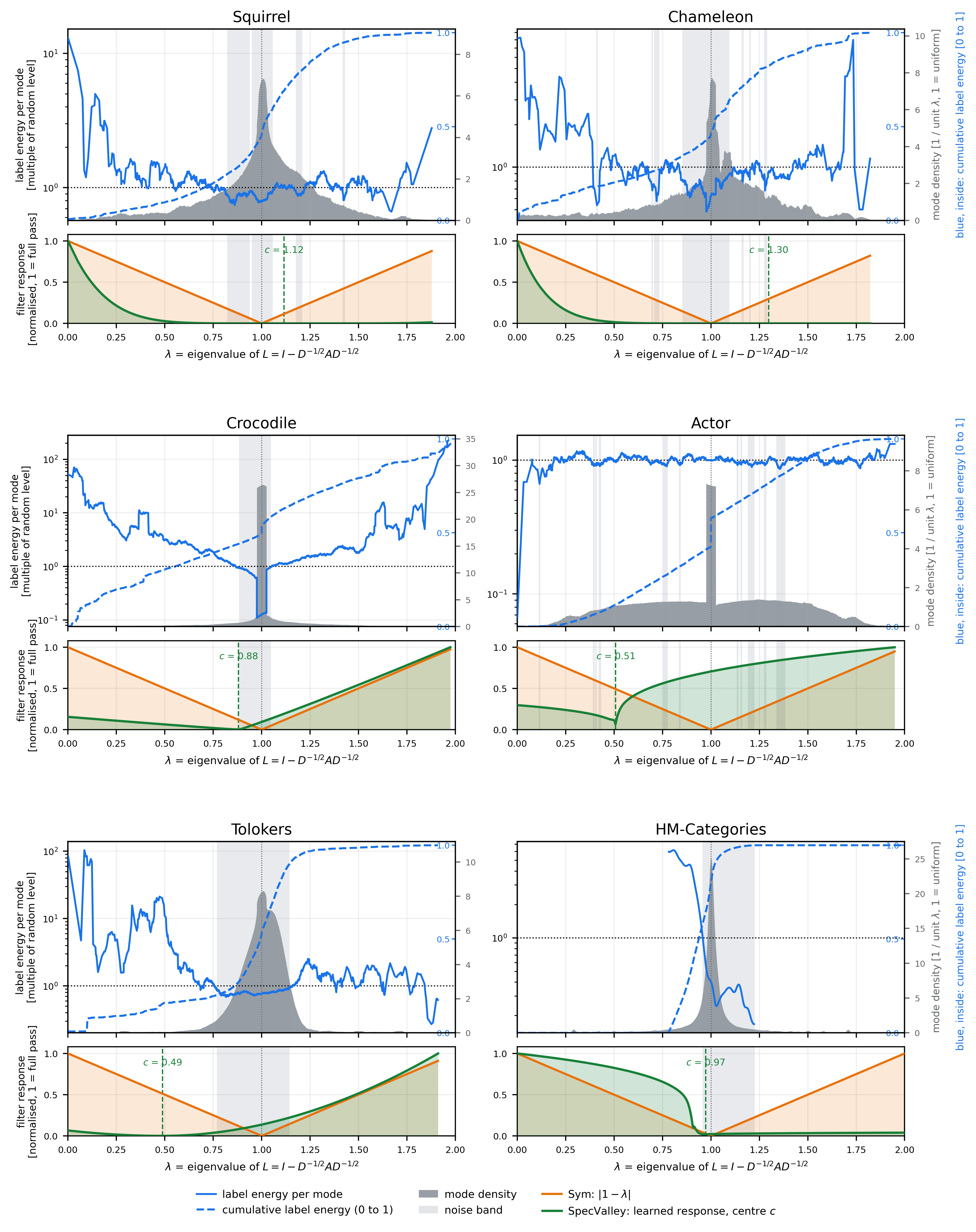}
    \caption{Label energy across the Laplacian spectrum (top) and magnitude of
    the fixed and learned aggregation responses (bottom) for all six graphs.}
    \label{fig:specvalley_app}
\end{figure}

Figure~\ref{fig:specvalley_app} extends Figure~\ref{fig:specvalley} to all six
graphs, with $\hat e_k$ as defined in
Section~\ref{subsec:spectral_properties_datasets} and $L = U\Lambda U^\top$ as
in Appendix~\ref{app:lemma_valley}. The solid curve reports $n\,\hat e_k$,
averaged over a sliding window of $m$ modes, so that $1$ is the level of a
labelling uncorrelated with the graph; the dashed curve is the cumulative share
$\sum_{\lambda_j\le\lambda}\hat e_j$. The grey area is the eigenvalue density,
normalised so that $1$ corresponds to the uniform distribution on $[0,2]$, and
the light band marks modes with $n\,\hat e_k < 1-z\sqrt{2/((C-1)m)}$, a
one-sided test at $z=1.645$ with $C$ the number of classes.

The lower panel shows what the two aggregators do with this spectrum. Degree normalisation acts

as the spectral function $1-\lambda$
(Proposition~\ref{prop:dense_valley}), a valley whose zero is pinned at
$\lambda=1$ by the normalisation, not by the data. SpecValley replaces it by the
learned response $r(\lambda)$ of Eq.~\eqref{eq:valley_response}, with centre $c$,
sharpness $p>0$ and flank gains $a_L, a_R$; training starts at $|1-\lambda|$
($c=p=1$, $a_L=a_R=1$). We plot the magnitude of each response normalised to its
maximum, which turns the zero into a visible valley and $|1-\lambda|$ into a V
with its vertex at $\lambda=1$. Two consequences of the normalisation should be
kept in mind: a large $p$ makes the response hug the axis over a wide range even
though it is not zero there, and a strong asymmetry $a_L\gg|a_R|$ flattens the
weaker flank relative to the stronger one.

\paragraph{Reading the six graphs.}
The density panels show the accumulation of Proposition~\ref{prop:dense_valley}:
on Crocodile, Tolokers and HM-categories $76\%$, $66\%$ and $79\%$ of the modes
lie in $|\lambda-1|<0.1$, on Squirrel and Chameleon the density forms a broader
hill ($41\%$, $40\%$), and on the sparser Actor graph ($27\%$) it is weakest.
The graphs differ in where the label energy lies relative to this core. On
Crocodile, Squirrel and Chameleon the core carries little of it ($17\%$ on
Crocodile), so $1-\lambda$ removes mainly modes at or below the random level. On
Tolokers the core carries $52\%$ of the label energy, so $1-\lambda$ removes
label information together with the noise. On HM-categories the label energy
concentrates in a narrow band just below $\lambda=1$ and falls below the random
level above it; on Actor it is flat at the random level. The learned responses
follow these differences: on Crocodile the valley stays close to $|1-\lambda|$,
on Tolokers its centre moves away from $\lambda=1$, and on HM-categories it
passes this band just below $\lambda=1$ and strongly attenuates everything
above it. The figure is descriptive: it shows where the label energy lies and
how the learned response places itself relative to it, but no single spectral
statistic predicts the benefit of aggregation on all six graphs
(Section~\ref{subsec:spectral_properties_datasets}).

\subsection{Polynomial Order of the Downstream Filter}
\label{app:polynomial_order}

\begin{wrapfigure}{r}{0.55\textwidth}
\vspace{-10pt}
    \centering
    \includegraphics[width=\linewidth]{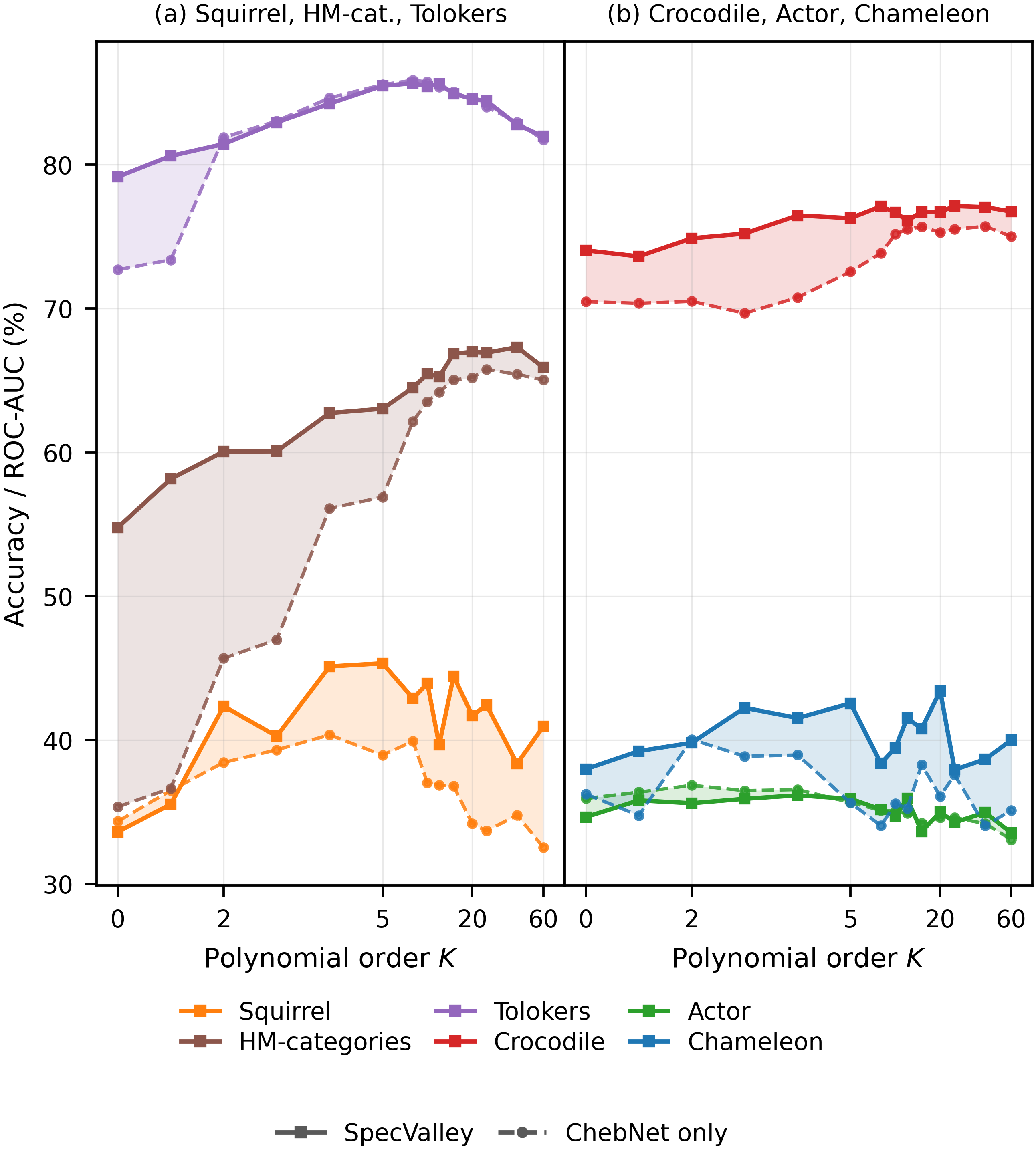}
    \caption{Polynomial order $K$ of the ChebNet layer, with and without the
    valley aggregator. Test score at the epoch with the best validation score,
    averaged over three splits (one on HM-categories), 150 epochs per order.}
    \label{fig:polynomial_order}
\vspace{-10pt}
\end{wrapfigure}

As discussed in Section~\ref{sec:sparsification}, we sweep the order
$K \in \{0,\dots,5\} \cup \{8,10,12,15,20,25,40,60\}$ of the downstream ChebNet layer
(Eq.~\eqref{eq:cheb_conv}), once with the valley aggregator of
Eq.~\eqref{eq:valley_operator} in front of it and once without, under the
protocol of Section~\ref{sec:experimental_settings} with 150 epochs per order.
Figure~\ref{fig:polynomial_order} shows the two curves per graph, grouped into
two panels for legibility.

On the graphs where the valley is effective (Squirrel-f, Chameleon-f,
Crocodile and HM-categories) the gap does not close with $K$: the filter
alone does not reproduce what the aggregator supplies, even at orders far beyond
those used in the benchmark. The composition of aggregator and filter is a
polynomial of only one order higher
(Section~\ref{subsec:Compatibility_Aggregation_Filter}), so this is not a
question of expressive capacity but of what the two parameterisations reach in
training. On Tolokers and Actor the two curves coincide from moderate orders
onward, consistent with a spectrum in which no aggregator finds a band to
attenuate (Section~\ref{subsec:spectral_properties_datasets}); there the
aggregator contributes only at very low orders, where the filter is too
constrained to use the graph at all.

\end{document}